\pdfoutput=1

\documentclass[11pt]{article}
\usepackage[utf8]{inputenc}
\DeclareUnicodeCharacter{03A9}{\ensuremath{\Omega}}

\usepackage[final]{acl}

\usepackage{times}
\usepackage{latexsym}

\usepackage[T1]{fontenc}
\usepackage[utf8]{inputenc}
\usepackage{inconsolata}

\usepackage{graphicx}
\usepackage{lipsum}

\usepackage[frozencache,cachedir=.]{minted}

\setminted{style=default}

\usepackage{hyperref}       %
\usepackage{url}            %
\usepackage{booktabs}       %
\usepackage{amsfonts}       %
\usepackage{nicefrac}       %
\usepackage{microtype}      %
\usepackage{pifont}

\usepackage{lipsum}
\usepackage{footnote}
\usepackage{todonotes}
\usepackage{colortbl}

\usepackage{xspace}
\usepackage{xparse}
\usepackage{graphicx}
\usepackage{amsmath}
\usepackage{array}

\usepackage[most]{tcolorbox}
\usepackage{xcolor}
\usepackage{fvextra}

\usepackage{makecell}  %

\usepackage{pdfpages}

\usepackage{listings} %

\lstdefinestyle{mystyle}{
    language=,               %
    basicstyle=\ttfamily\scriptsize, %
    keywordstyle=\color{blue},    %
    stringstyle=\color{red},      %
    commentstyle=\color{gray},    %
    showstringspaces=false,       %
    breaklines=true,              %
    breakatwhitespace=false,      %
    prebreak=\textbackslash,      %
    postbreak=\mbox{\textcolor{gray}{$\hookrightarrow$}\space}, %
    upquote=true,                 %
    columns=fullflexible,         %
}

\definecolor{ideationred}{RGB}{255, 89, 94} %
\definecolor{planningyellow}{RGB}{255, 202, 58} %
\definecolor{codinggreen}{RGB}{138, 201, 58} %
\definecolor{reportingblue}{RGB}{25, 130, 196}
\definecolor{metaanalyspurple}{RGB}{106, 76, 147}

\NewDocumentCommand{\rot}{O{45} O{1em} m}{\makebox[#2][l]{\rotatebox{#1}{#3}}}%
\NewDocumentCommand{\rotb}{O{-45} O{1em} m}{\makebox[#2][l]{\rotatebox{#1}{#3}}}%

\newcommand{\eat}[1]{}

\usepackage{quoting}

\newcommand{\codesci}{\textsc{CodeScientist}\xspace}

\title{\raisebox{-0.3\height}{\includegraphics[height=1cm]{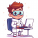}}
\codesci : End-to-End Semi-Automated Scientific Discovery\\ with Code-based Experimentation}

\author{
 \textbf{Peter Jansen\textsuperscript{1,3}},
 \textbf{Oyvind Tafjord\textsuperscript{1}},
 \textbf{Marissa Radensky\textsuperscript{1}},
 \textbf{Pao Siangliulue\textsuperscript{1}},
\\
 \textbf{Tom Hope\textsuperscript{1,4}},
 \textbf{Bhavana Dalvi Mishra \textsuperscript{1}},
 \textbf{Bodhisattwa Prasad Majumder\textsuperscript{1}},
\\
 \textbf{Daniel S. Weld\textsuperscript{1,2}},
 \textbf{Peter Clark\textsuperscript{1}}
\\
\\
 \textsuperscript{1}Allen Institute for Artificial Intelligence,
 \textsuperscript{2}University of Washington,\\
 \textsuperscript{3}University of Arizona,
 \textsuperscript{4}Hebrew University of Jerusalem 
\\
 \small{
   \texttt{peterj@allenai.org}
 }
}

\begin{document}
\maketitle
\begin{abstract}
Despite the surge of interest in autonomous scientific discovery (ASD) of software artifacts (e.g., improved ML algorithms), current ASD systems face two key limitations: (1) they largely explore variants of existing codebases or similarly constrained design spaces, and (2) they produce large volumes of research artifacts (such as automatically generated papers and code) that are typically evaluated using conference-style paper review with limited evaluation of code. In this work we introduce \codesci, a novel ASD system that frames ideation and experiment construction as a form of genetic search jointly over combinations of research articles and codeblocks defining common actions in a domain (like prompting a language model).  We use this paradigm to conduct hundreds of automated experiments on machine-generated ideas broadly in the domain of \textit{agents and virtual environments}, with the system returning 19 discoveries, 6 of which were judged as being both at least minimally sound and incrementally novel after a multi-faceted evaluation beyond that typically conducted in prior work, including external (conference-style) review, code review, and replication attempts. Moreover, the discoveries span new tasks, agents, metrics, and data, suggesting a qualitative shift from benchmark optimization to broader discoveries.\footnote{\url{https://github.com/allenai/codescientist}}

\end{abstract}

\section{Introduction}

Automated scientific discovery (ASD) systems have already had success in targeted domains like protein folding (\textsc{AlphaFold}, \citealp{alphafold}), antibiotic discovery \cite{halicin}, and model optimization (\textsc{Lion}, \citealp{lion}), by using custom \textit{problem-specific} systems that search (large) hand-crafted search spaces. 
Recently, language models (LMs) are fueling explorations into more \textit{problem-general} discovery systems capable of the full research pipeline of ideation, planning, (code-based) experimentation, and experiment analysis, with numerous impressive systems appearing recently, including \textsc{AI Scientist} \cite{aiscientist}, \textsc{AIGS} \cite{aigs}, \textsc{AgentLab} \cite{agentlab}, and \textsc{Data-to-Paper} \cite{DataToPaper}.  Impressive as these systems are, each makes simplifications to reduce complexity, such as restricting search to variants of prewritten code, using a DSL for experiments, or working on restricted problems. %

In this work we introduce \codesci, an ASD system built with novel innovations for ideation and experiment execution -- incorporating genetic search \cite{llm-gp} over combinations of literature and code -- that we hypothesize will increase the diversity of the discoveries the system makes.  We run our system at scale (hundreds of experiments) in the broad domain of \textit{agents and virtual environments}, and find that of the 19 discoveries our system suggests, 6 appear to meet minimum thresholds for scientific soundness and incremental novelty after domain expert review.  Moreover, the discoveries our system produces qualitatively appear diverse, and span creating new tasks, agents, metrics, data, and challenging assumptions, which builds-upon (while broadening) the scope of the impressive accomplishments of existing systems that focus on improving model performance on standardized ML benchmarks \citep[e.g.][]{agentlab,Li2024MLRCopilotAM,mlagentbench}. %

\begin{figure*}[t]
  \centering
  \includegraphics[scale=1.05]{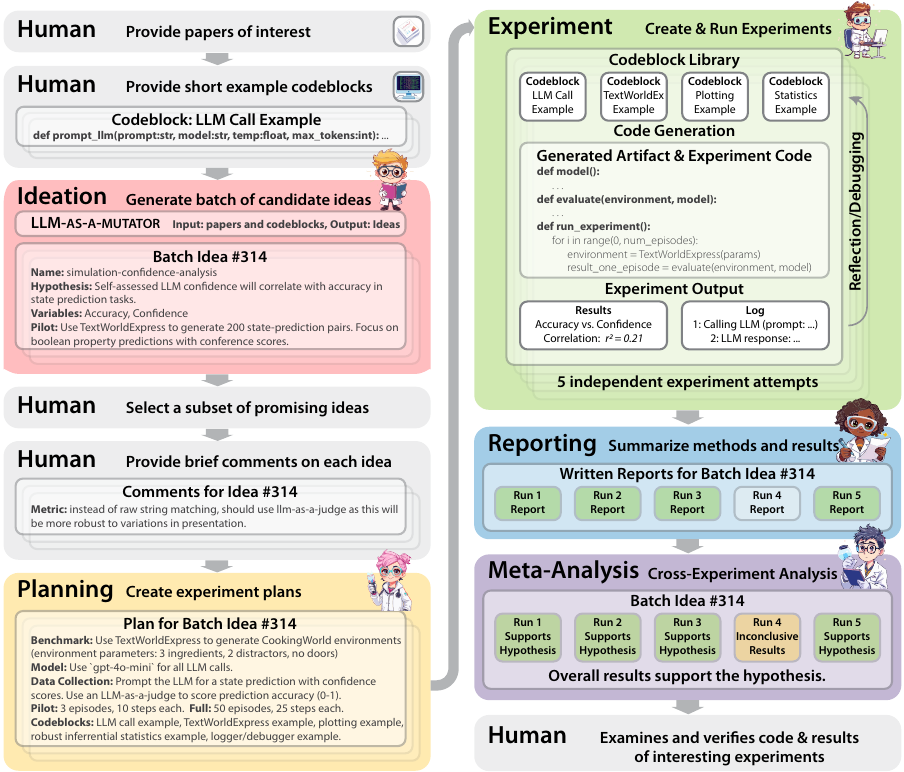}
  \caption{\footnotesize An overview of the core stages of the \codesci discovery workflow, including ideation, planning, building and executing code-based experiments, reporting results, and performing meta-analyses across experiments.}
  \label{fig:system_overview}
\end{figure*}

Orthogonally, ASD research itself is faced with significant methodological challenges that limit progress.
The first, \textbf{evaluating discoveries}, is complex in that scientific discovery is (by definition) at the edge of human knowledge, and as such, tasks with gold annotated outcomes \citep[e.g., \textsc{SWE-Bench}, ][]{swebench} are generally unavailable. Alternative assessments are needed, for example the way humans evaluate research -- namely, rigorous (and expensive) manual review, or the use of proxy metrics \cite{aiscientist} that are progressively improving their agreement with human ratings \cite{Radensky2024ScideatorHS}. The second challenge, \textbf{variability}, acknowledges that even when using low temperatures, workflows built upon language models rarely produce the same output -- especially when each step autoregressively depends upon the output of the previous step.  Across successive runs, the specific ideas that an ASD system generates are different, and (as we show in this work), the code it implements for those ideas (and whether it succeeds or fails) is also highly variable -- making comparisons across systems (or ablations of a single system) costly, methodologically challenging, and uncommon in the literature.  At the same time, with ASD still in its infancy, scientifically sound and novel discoveries are relatively rare -- and (like other code-generation tasks like \textsc{SWE-Bench}), systems are faced with high variability and low absolute success, but (unlike \textsc{SWE-Bench}) without the benefit of trustworthy automatic evaluation. %
In spite of these methodological challenges, we show that \codesci is capable of generating a number of candidate discoveries through manual (domain-expert) evaluation, and that this set of discoveries qualitatively captures a series of diverse research ideas that expand the scope of accomplishments of existing systems.

\begin{table*}[t!]
\centering
\footnotesize
{\setlength{\tabcolsep}{3pt}
\begin{tabular}{lccccccc}
\toprule
\textbf{}                                        &   \textbf{Ideation}              & \textbf{Artifact}   &   \textbf{Evaluation}    &   \textbf{Evaluation} \\%
\textbf{Discovery System}                        &   \textbf{Methodology}           & \textbf{Evaluated}  &   \textbf{(Automatic)}   &   \textbf{(Human)}      \\%
\midrule
\rowcolor[HTML]{F3F3F3} 
\textsc{AIScientist}~\citep{Lu2024TheAS}         &   mutate existing experiment &  paper            & Likert (NeurIPS)      &   No  \\
\textsc{AIGS}~\citep{aigs}                       &   task + past experiments    &  paper            & Natural Language      &   Likert   \\
\rowcolor[HTML]{F3F3F3} 
\textsc{AgentLab}~\citep{agentlab}               &   human + literature         &  benchmark        & Likert                &   Likert      \\
\textsc{Data-to-Paper}~\citep{DataToPaper}       &   table analysis             &  code             & Rules                 &   Code Review \\ 
\rowcolor[HTML]{F3F3F3} 
\textsc{MLR-Copilot}~\citep{Li2024MLRCopilotAM}  &   literature                 &  benchmark        & Likert                &   Likert \\
\rowcolor[HTML]{E3E3E3} 
\textbf{\codesci (this work)}                    &   \textbf{literature + codeblocks}    &   \textbf{paper+code}            & \textbf{Accept/Reject Hyp.}    &  \makecell{\textbf{Likert (paper)+}\\ \textbf{Code Review}} \\
\bottomrule        
\end{tabular}
}
\caption{\footnotesize A comparison of existing discovery systems with \codesci, in terms of their ideation methodology, which research artifact is evaluated (paper, code, or performance on a benchmark), and which automatic or manual evaluations are performed on the research results.\label{tab:env_comparison}}
\end{table*}

The contributions of this work are:
\begin{enumerate}
\item \codesci, a novel, open-source, end-to-end system for semi-automated scientific discovery, which ideates and executes experiments based on genetic search over both literature and a library of codeblocks. 
\item A demonstration in the domain of \textit{agents and virtual environments}, where we show that \codesci discovers 6 incremental yet novel research results not seen previously.  These results were validated by external reviewers (in a conference-style review), then further vetted by replication and code review.
\item A qualitative analysis of failure modes and research methods challenges, to help guide future research on ASD technology. This includes an examination of the benefits of incorporating human input into our workflow, versus using a completely automated system.
\end{enumerate}

\section{Related Work}

{\flushleft\textbf{Ideating and Executing Research:}} The suite of recent ASD systems primarily differ in their methods for ideation, experiment construction and execution, as well as their problem domain (e.g. chemistry, biology, AI) -- with examples of systems in the AI domain provided in Table~\ref{tab:env_comparison}. Ideating on literature is a common (and naturalistic) method, but unrestricted, can produce broadly scoped ideas that are challenging to implement.  Ideating on existing experiment code, essentially applying the LLM as a mutation operator \cite{llm-gp} to hypothesize how code changes might (for example) increase performance on a benchmark, helps narrow ideas to those likely to be executable.  Similarly, one can restrict ideation to problem-specific templates for specific tasks \cite{aigs, DataToPaper}.  In this work, we explore a novel genetic search ideation strategy that combines two facets: the open-endedness of literature-based ideation, with a focus on generating ideas that are implementable by our experiment builder by partially conditioning the ideation on a library of codeblocks available to the executor that implement common research functions (like calling a language model, or creating a plot). 

{\flushleft\textbf{Evaluating Discoveries:}} 
Automatic evaluation of discoveries is possible in a limited subset of domains and tasks -- such as in materials science, where molecular simulators can directly evaluate the veracity of discoveries \cite{Qi2024MetaScientistAH}. Similarly,  broader science-themed games can include instrumentation for measuring discoveries \cite{discoveryworld}, or the scope of discoveries can be reduced to some measurable quantity, such as increasing model performance on a benchmark task \cite{mlagentbench}.  Manual evaluation of research artifacts (such as code) is costly to perform, and previously limited in scope \citep[e.g., statistical analysis code in][]{DataToPaper}. Others have worked to develop automated proxy metrics that use an \textsc{LLM-as-a-Judge} paradigm to evaluate research papers on Likert scales \cite{likert1932technique} similar to conference review \cite{aiscientist, agentlab, Li2024MLRCopilotAM}, though due to the difficulty of this task, these currently may have limited agreement with human judgements \cite{Radensky2024ScideatorHS}, may be fooled by superficial niceities (fluency, layout, etc.), and (as in the case of \textsc{AI Scientist}), may reject all the discoveries.\footnote{\textsc{AI Scientist} recently announced that it submitted 3 AI-generated papers from an unreleased (\textsc{v2}) system to a workshop, one of which was accepted: \url{https://sakana.ai/ai-scientist-first-publication-jp/}; Similarly intology.ai recently reported two AI-generated papers accepted for workshops \cite{zochi}.}  As we show in Section~\ref{sec:discovery-experiments}, faithfulness is a critical factor to automated evaluation --  what our system says it has discovered in its papers can strongly differ from what it actually implements in code, and we argue that conference-style Likert assessments of papers (without code) may not fully assess the faithfulness of science-as-code discoveries.

\section{System Overview}

An overview of the \codesci system is shown in Figure~\ref{fig:system_overview}, with the 5 major steps of the workflow (\textit{ideation}, \textit{planning}, \textit{experiment building}, \textit{reporting}, and \textit{meta-analysis}) described below. Except where described otherwise, the steps of the workflow are implemented as prompts to a language model, with example prompts and additional implementation details provided in \textsc{Appendix}~\ref{sec:appendix-prompts}.

\subsection{Paper Corpus and Codeblocks}
\codesci requires two forms of pre-generated input: (a) a human-curated list of papers to ideate from in the users domain(s) of interest, and (b) a set of relevant codeblocks that demonstrate how to perform common tasks. Example codeblocks include \textit{how to call an LLM}, \textit{how to implement a ReAct agent}, and \textit{how to load specific benchmarks}.  More on the specific agent-centered corpus of papers and codeblocks used in this work is provided in Section~\ref{sec:discovery-experiments}.

\subsection{Ideation}
The purpose of the ideator is to generate a large set of candidate research ideas (conditioned on recent research articles) that could be explored by \codesci.  Pilot studies showed that nearly all of the ideas generated by our early ideator were either too complex or too open-ended to be implemented by our automated experiment building system, such as requiring the ability to download and modify arbitrary models and benchmarks mentioned in the input papers.  To generate ideas that are both conditioned on the literature and on code components that have a high chance of successfully executing in the execution sandbox, the ideator prompt includes both (a) two randomly chosen papers, from a corpus of papers provided by a human scientist, and (b) summaries of codeblocks that form a vetted library of common research-related functions.    More formally, the ideator takes a human-generated corpus of papers ($papers$) and a library of codeblocks ($codeblocks$) as input, producing a set of candidate ideas ($ideas$) as output: 
\begin{equation}
ideas = \operatorname{Ideator}(papers, codeblocks)
\end{equation}

\noindent Each idea ($i \in ideas$) is structured with slots for the \textit{hypothesis}, \textit{dependent/independent variables}, \textit{evaluation metrics}, \textit{baselines}, \textit{pilot experiment design}, and a list of \textit{major sections of code and other resources} anticipated to be required to successfully implement the idea, with slot values populated in natural language. %
To emphasize creating diverse combinations of ideas conditioned on ideas in the literature while using genetic-style search \cite{llm-gp}, the prompt includes sample types of ideation to serve as genetic operators, including cross-over (i.e. \textit{combining ideas}), and mutation (i.e. \textit{extending ideas}, \textit{challenging assumptions}, \textit{filling in gaps}, and so forth).
After generating a large pool of candidate ideas, as a cost-saving and efficiency measure, a human then manually selects a subset of these ideas, $ideas_f \subset ideas$, that appear most interesting to them.  For each idea $i \in ideas_f$, the domain expert can also provide a brief (2-3 sentence) set of human comments $h$, such as suggesting alternate metrics or benchmarks to use, that increase the utility and tractability of the idea. An example idea, human comment, and plan is provided in \textsc{Appendix}~\ref{sec:idea-operationalization-example}.\footnote{The full set of human comments is provided in \textsc{Appendix}~\ref{sec:domain-expert-comments}, and an ablation in the discussion suggests removing expert comments reduces the number of discoveries judged as minimally sound and having at least incremental novelty by approximately one third.}

\subsection{Planning}

The planning step converts the high-level idea generated by the ideator into a more detailed, practical, and operational experiment plan for the experiment builder.  Where the ideator generates high-level ideas such as \textit{``examine whether a ReAct agent augmented with a causal memory increases performance on benchmark X''}, the planning step generates a specific plan for implementing the \textit{artifact} (i.e. the augmented \textsc{ReAct} agent) %
as well as the \textit{experiment} to test its properties (such as which base model to use, hyperparameters, and other experimental details).  More formally, the $\operatorname{Planner}$ takes as input an idea $i$, expert comments on that idea $h$, and the codeblock library $C$ as input, producing a plan $p$ and anticipated list of codeblocks $c \subset C$ required to implement that plan as output:
\begin{equation}
    {p, c} = \operatorname{Planner}(i, h, C)
\end{equation}

\noindent This experiment plan and list of codeblocks serve as input to the experiment construction system.

\subsection{Experiment Construction and Execution}

The experiment builder is tasked with generating the code for the \textit{artifact} and \textit{experiment} through an iterative series of \textit{generate-execute-reflect} debugging steps.  The builder takes an experiment plan $p$ and list of codeblocks $c$ as input, and produces a set of generated code $g$, experimental results $r$, and output logs $l$ as output:
\begin{equation}
    g, r, l = \operatorname{Builder}(p, c)
\end{equation}

\noindent The experiment builder includes three tightly-coupled components:
    {\flushleft\textbf{Initial code generation:}} Synthesizes \textsc{Python} code to implement the artifact and experiment based on the plan and selected codeblocks. 
    {\flushleft\textbf{Instrumented execution sandbox:}} The generated code is executed within an instrumented sandbox. This sandbox captures the full output of the code -- including logging and standard output/error streams, and intercepts \textsc{API} calls (e.g. to \textsc{OpenAI}, \textsc{Anthropic}, or \textsc{Together.ai} models) via a proxy that tracks usage statistics and enforces cost limits.
    {\flushleft\textbf{Reflection and debugging:}} At each debug iteration, the model is asked to reflect on the code and output (logs, streams, and \textsc{API} usage statistics), and determine if the experiment has completed successfully and faithfully.  If successful, the experiment building step concludes and progresses to reporting.  If unsuccessful, the reflection step modifies the code, and the \textit{execute-reflect} cycle continues until the experiment is marked as complete, or a hard limit (time, cost, or number of debug iterations) is reached.

For pragmatic reasons (cost, runtime, prompt length), the experiment builder first attempts to build a short pilot experiment with inexpensive and fast debug cycles.%
When the experiment builder determines the pilot experiment has succesfully completed, it automatically scales to running (and, if necessary, further debugging) the full experiment. %

\begin{table}[t!]
\centering
\scriptsize
\begin{tabular}{p{0.08\linewidth} p{0.80\linewidth}}
\toprule
\textbf{Rating} & \textbf{Automated Result Summary} \\
\midrule
\rowcolor{green!20}
supports    & Discovered action templates significantly improved agent performance in CookingWorld, outperforming both baseline and manual templates.\\
\rowcolor{yellow!20}
inconcl. & Decomposition history slightly improved agent performance (0.183 vs 0.117) but results weren't statistically significant. \\
\rowcolor{red!20}
reject  &   Knowledge graph-based mode switching showed no clear advantage over baseline strategies in ScienceWorld tasks.\\
\bottomrule
\end{tabular}
\caption{\footnotesize Example result summaries and corresponding ratings for whether they \textit{support, reject}, or are \textit{inconclusive} towards the respective experiment hypothesis.  Each of the 3 examples comes from a different idea. \label{tab:result-summary-examples}}
\end{table}

\subsection{Reporting}
Successfully completed experiments enter an automated reporting step that takes the experiment plan ($p$), code ($g$), results ($r$), and logs ($l$) as input and produces both a written \LaTeX~report $w$, and a short summary report $s$ as output:
\begin{equation}
    w, s = \operatorname{Reporter}(p, g, r, l)
\end{equation}

\noindent Because the system can have a high throughput and generate experimental results faster than a human could read the detailed reports ($w$), the short high-level summaries of results ($s$) help a user determine if reading the full report is warranted by highlighting the main results.  This includes explicitly categorizing whether the hypothesis of the experiment was \textit{confirmed} or \textit{rejected} by the experimental results, or if the results were \textit{inconclusive}. Example summaries and ratings are shown in Table~\ref{tab:result-summary-examples}.

\subsection{Meta-Analysis}
Pragmatically, even when provided with the same plan, codeblocks, and a low generation temperature, the experiment builder frequently produces different implementations across successive runs due to the inherent variability introduced both by the language models themselves, as well as by autoregressively conditioning the code they generate on the output of the previous debug cycle.  At the same time, language models still struggle with many complex scientific code generation tasks, and some experiments may fail to create a successful implementation.  To reduce this variability, we include a meta-analysis step where for each idea and plan, the experiment builder is independently run $N$ times, producing $N$ different experimental results.  The meta-analysis then examines the consistency of the results across successive experimental implementations, generating a meta-analysis report $m$: 
\begin{equation}
    m_i = \operatorname{MetaAnalysis}({s_1, s_2, s_3, ..., s_N})
\end{equation}

\noindent where $s_n$ represents a single result summary from running a given plan $p$ through the experiment builder and reporting process $N$ times.

\section{Discovery Experiments}
\label{sec:discovery-experiments}
Candidate discoveries made by \codesci are described in Section~\ref{sec:successful-discoveries}, with challenges and failure modes described in Section~\ref{sec:failures-and-challenges}.  A description of the experiment setup is provided below.\footnote{Unless otherwise stated, we use \textsc{Claude-Sonnet-3.5-1022} as our base model for \codesci. We ask the planner to prefer to experiment on \textsc{GPT-4o-mini}, due to its high speed and low cost.}

{\flushleft\textbf{Paper and Code Example Corpus:}} We assembled a corpus of 57 recent papers broadly in the area of agent architectures and virtual environments.  Paired with this, we assembled 10 example code snippets\footnote{Code snippets were assembled from a combination of existing library examples and documentation, LLM-generation, and/or manual authoring.} for performing basic tasks in this domain, including calling language models \cite{openai2024gpt4technicalreport}, building a ReAct agent \cite{yaoreact}, plotting \cite{hunter2007matplotlib}, robust inferential statistics for comparing models \cite{efron1992bootstrap}, using common knowledge graphs \cite{speer2017conceptnet,miller1995wordnet}, and several benchmark environments including \textsc{TextWorldExpress} \cite{jansen-cote-2023-textworldexpress}, a high-performance simulator that reimplements common benchmarks.  These code examples serve not only as vetted code examples of common artifacts described in the paper corpus, but also as demonstrations of the nuances required to implement code within the sandbox environment. 

{\flushleft\textbf{Idea Selection:}} {We used \codesci to generate approximately 2000 candidate experiment ideas ideated from 200 randomly selected combinations of papers. As a cost saving measure, a stratified sample was presented to a domain expert, until they had manually selected 50 ideas that appeared both viable and sufficiently different from one another. Each idea was provided with brief comments by the domain expert (included in \textsc{Appendix}~\ref{sec:domain-expert-comments}, to show the minor corrections these entail), then plans for each idea were generated.

\begin{table}
\centering
\footnotesize
    \begin{tabular}{lc}
    \toprule
    \textbf{Configuration}   \\
    \midrule
    ~~Number of ideas evaluated               &   50  \\
    ~~Experiment Builder attempts per idea    &   5   \\
    ~~Total experiment attempts               &   250 \\
    \midrule
    \textbf{Enforced Limits (per experiment)}    \\
    \midrule
    ~~Maximum Debug Iterations                &   25  \\
    ~~Total cost limit                        &   \$10    \\
    ~~LLM cost limit (per debug iteration)    &   \$1 \\
    ~~Execution time limit (per debug iteration)     &   90 min.  \\
    ~~Hard time limit (all debug iterations)  &   6 hours \\
    \midrule
    \textbf{Average Usage (per experiment; N=250)}   \\
    \midrule
    ~~Average debug iterations        &   15.8  \\
    ~~Average cost                    &   \$4.23  \\
    ~~Average runtime                 &   131 min.  \\    
    ~~Average generated code length (lines)    &   506    \\
    ~~Average generated code length (tokens)     &   4521   \\
    \bottomrule
    \end{tabular}
    \caption{Summary statistics for discovery experiments.\label{tab:summary-statistics}}    
\end{table}

\begin{table*}[t!]
\centering
\footnotesize
\resizebox{\linewidth}{!}{%
\begin{tabular}{>{\centering\arraybackslash}p{0.02\linewidth} >{\centering\arraybackslash}p{0.07\linewidth} >{\centering\arraybackslash}p{0.07\linewidth} p{0.82\linewidth}}
\toprule
\textbf{} & \multicolumn{2}{c}{\textbf{Human Reviewers}} &  \textbf{} \\
\textbf{} & \textbf{Min.} & \textbf{Some} &  \textbf{} \\
\textbf{\#} & \textbf{Sound}    & \textbf{Novelty}     &  \textbf{Description of Discovery} \\
\midrule
\multicolumn{4}{l}{\textbf{Candidate discoveries identified by external reviewers, and supported by internal review}}    \\
\midrule
\rowcolor[HTML]{F3F3F3} 
1   &   1.0 & 1.0  &   \textbf{State Prediction Confidence:} In a state prediction task, an LLM's self-assessed confidence in its predictions have a low corelation with the accuracy of those predictions. (The state prediction data was automatically crawled from one of the benchmarks) \textit{(Though the correlation varies across experiments, the value consistently appears low.)} \\%
2   &   1.0 & 1.0 &   \textbf{Accuracy vs Representational Expressivity:} In a state prediction task, an LLM performs better at predicting simpler representations (e.g. boolean values) versus states including text. (The state prediction data was automatically crawled from one of the benchmarks) \textit{(Significant implementation and evaluation differences across experiments, but generally support the idea that predicting simpler representations is easier.)} \\%
\rowcolor[HTML]{F3F3F3} 
3   &   1.0 & 1.0 &   \textbf{Multi-Stage Environment Generation:} When creating novel benchmark environments using code-generation, generating the environments in multiple stages increases environment fidelity. \textit{(A small change on LLM-for-environment-generation tasks, implementing specific aspects in each step, rather than generating as a whole and reflecting. For evaluation, creates a simple proxy metric that seems well-motivated as this type of evaluation is an open problem in the literature, and even llm-as-a-judge paradigms have issues with this task, while being vastly more expensive).} \\ %

4   &   1.0 & 1.0 &   \textbf{Combinatorial Optimization:} A language model performs poorly at a combinatorial optimization problem (selecting values from a set that are closest to adding to a specified value X), grounded in substituting resistor values in electronics. \textit{(Consistent result across experiments, and tested to within different tolerances, e.g. 1\%, 5\%)} \\ %
\rowcolor[HTML]{F3F3F3}
5   &  1.0 & 0.66 &   \textbf{Action Prediction:} An LLM's ability to predict whether actions will be successful in a virtual environment is generally low, marginally above a random baseline. \textit{(Appears true, with the following qualifications: (1) the LLM was given only the current observation, and no history, to judge from, and (2) an LLM-as-a-judge was used to help collect the gold dataset, and has imperfect labels.)} \\ %
6   &   0.66 & 0.66 &   \textbf{Graph Agent for Discovery:} A ReAct agent augmented with a graph-based memory outperforms a ReAct baseline on a highly complex environment (\textsc{DiscoveryWorld}). \textit{(Appears true. Graphs appear relatively simple, forming a form of \textsc{object-property} memory, rather than complex nested relationships. )} \\ %

\midrule
\multicolumn{4}{l}{\textbf{Results rejected by either external reviewers, internal review, or both}}\\
\midrule
\rowcolor[HTML]{F3F3F3} 
7   &   1.0 & 1.0 &   \textbf{Social Graphs:} For a task in managing social relationships, an agent that uses a graph to keep track of those relationships does not outperform perform a simpler non-graph baseline. \textit{(The experiment-generated benchmark is overly simple, contains few samples, and the llm-as-a-judge metric is not validated)} \\ %
8   &   1.0 & 1.0 &   \textbf{Planning Agent:} A custom planning agent outperforms a ReAct agent on the CookingWorld benchmark. \textit{(Effect fails to replicate when number of samples increased)}.\\ %
\rowcolor[HTML]{F3F3F3}
9   &   1.0 & 1.0 &   \textbf{Spatial Agent:} A ReAct agent that maintains an explicit graph of interconnected locations outperforms a baseline ReAct agent. \textit{(Effect fails to replicate when number of samples increased)}.\\    %
10   &   1.0 & 1.0 &   \textbf{Action History:} A ReAct agent that keeps a history of recent actions (as well as whether those actions increased the task score) outperforms a ReAct baseline. \textit{(Effect fails to replicate when samples increased).}\\%
\rowcolor[HTML]{F3F3F3}
11   &   1.0 & 0.66 &   \textbf{Template Agent:} Creates an agent that applies templates of action sequences learned in the training set. \textit{(A non-LLM agent that is partly hard-coded, and partly reinventing a Markov model).}\\    %
12   &   0.66 & 1.0 &   \textbf{Graph Verification Agent:} An agent that builds and explicitly verifies its graph before using it outperforms a baseline agent without this verification step. \textit{(On examining the code, the experiment is incorrect because neither baseline nor experimental agents ever actually use the graph it builds, they just \textit{randomly} pick actions. An interesting secondary result -- that an LLM used to extract triples from an environment observation only verifies $\approx$80\% of those triples as correct in a secondary verification step -- does appear to be supported.)} \\ %
\rowcolor[HTML]{F3F3F3}
13   &   0.66 & 0.66 &   \textbf{Affordance Agent:} An LLM agent that predicts object affordances outperforms a random baseline on ScienceWorld. \textit{(Baseline is too simple)}\\ %
14   &   0.33 & 0.66 &   \textbf{WordNet:} A hardcoded WordNet agent outperforms a random baseline. \textit{(Baseline is too simple)}\\ %
\rowcolor[HTML]{F3F3F3}
15   &   0.33 & 0.66 &   \textbf{Metaphor Graphs:} Examines generating graphs that describe the metaphorical relationships between objects in virtual environments. \textit{(Unclear motivation and utility)}\\ %
16  &   0.0 & 1.0 &   \textbf{Goal Tracking:} A ReAct agent augmented with goal tracking outperforms a random baseline. \textit{(Baseline is too simple)}\\ %
\rowcolor[HTML]{F3F3F3}
17   &   0.0 & 1.0 &   \textbf{Container Agent:} An agent built specifically to handle objects inside containers (a challenge for some agents) outperforms a random baseline. \textit{(Baseline is too simple)}\\ %
18   &   0.0 & 0.66 &   \textbf{Template-based Environment Creation:} Parametrically generating new environment benchmarks using templates to define the environment is just as good as manually building the environments. \textit{(Both the template and ``manual'' (should be human-generated?) environments were hardcoded in the experiment code -- so this was a foregone conclusion).} \\%
\rowcolor[HTML]{F3F3F3}
19   &   0.0 & 0.66 &   \textbf{Graph Metric:} A custom-developed metric for determining similarity between text and graph-based representations of environments outperforms Jaccard similarity. \textit{(Report lacks details to evaluate.)} \\%
\bottomrule
\end{tabular}%
}
\caption{\footnotesize Descriptions of the 19 discoveries \codesci automatically identifies from 250 experiments over 50 ideas. \textit{Human Reviewers} refers to average ratings from 3 external reviewers.  The 6 discoveries under ``candidate discoveries'' were rated as meeting minimum soundness and and incremental novelty criteria by external reviewers (examining papers), and similarly by the internal reviewer (examining both code and papers). The 11 discoveries under ``rejected`` received low ratings from the external reviewers, or received high ratings from external reviewers but were rejected by the internal reviewer either after examining the code and identifying issues with soundness, or failing to replicate results after rerunning experiments with a higher numbers of samples. Comments from the internal reviewer are provided in \textit{italics}. \label{tab:discoveries} }%
\end{table*}

{\flushleft\textbf{Experiment Builder:}} For each of the 50 ideas, we called the experiment builder 5 times (i.e. each idea was given 5 attempts to generate functioning experiment code and generate results), for a total of 250 experiment runs.  Cost, runtime limits, and actual usage statistics are provided in Table~\ref{tab:summary-statistics}.  %

{\flushleft\textbf{Reporting and Meta-Analysis:}} 
The reporting stage produced written reports for each experiment (intended for the user, and provided in \textsc{Appendix}~\ref{sec:experiment-reports}), as well as short summaries of results.  Short summaries of each of the 5 experiments for a given idea were used for meta-analysis, which described whether the results across experiments generally support or reject the hypothesis.  Additional details of the meta-analysis procedure are provided in \textsc{Appendix}~\ref{sec:meta-analysis-categorization-criteria}.

\subsection{Candidate Discoveries}
\label{sec:successful-discoveries}
\codesci flagged 19 of 50 ideas for human inspection where at least one of the five experiment runs produced ``interesting'' results,\footnote{Results were flagged for humans using a heuristic prompt found in \textsc{Appendix}~\ref{sec:appendix-prompts}} with those discoveries provided in Table~\ref{tab:discoveries}.  %
We performed two separate evaluations on these candidate discoveries: external, and internal.  
{\flushleft\textbf{External (conference-style) Review:}} A conference style review. We recruited 3 external reviewers that are practicing research scientists in natural language processing and who have previously published in the domain (agents and virtual environments).  Reviewers were provided with \codesci generated papers, and asked to rate them on \textit{soundness} and \textit{novelty}, with the rubric provided in \textsc{Appendix}~\ref{sec:external-reviewer-rubric}.  Each domain expert rating was then converted to a binary score.  Ratings of \textit{unsound} or \textit{not novel} experiments were considered failures, and converted to zeros.  Ratings of \textit{clearly sound}, \textit{likely sound}, or \textit{minor concerns (not altering overall conclusions)} were considered meeting the minimum threshold for soundness, and converted to scores of one.  Similarly, ratings of \textit{incremental novelty} or above were considered meeting the minimum threshold for novelty, and converted to one.  The average of these binarized ratings across the three external reviewers is provided in Table~\ref{tab:discoveries}.  If the majority of reviewers rated a discovery as meeting minimum soundness and novelty thresholds, we considered it as having passed external review.

{\flushleft\textbf{Internal Review:}} A domain expert (one of the authors) provided an in-depth review of the code and experiment logs, and attempted to replicate the results with a larger number of samples.  This reviewer essentially functioned as a ``veto'', able to reject discoveries that appeared genuine from the paper and external review, but did not pass detailed examination.

{\flushleft\textbf{Candidate discoveries:}} Of the 19 candidate discoveries flagged by \codesci, 13 (68\%) were rated as meeting minimum soundness and novelty criteria by at least 2 of the 3 external reviewers.  When examining the code, experiment logs, and performing replication attempts, the internal reviewer rejected 7 of these 13 discoveries, resulting in a total of 6 discoveries (32\%) that passed both external and internal review. 

The discoveries passing these tests take a variety of forms. While some take the form of improving model performance on benchmarks, %
most involve creating new tasks, benchmarks, metrics, methods, or questioning assumptions.  The discoveries include determining that an LLM's self-assessed confidence in its prediction accuracy has a low correlation with its actual accuracy in state-prediction tasks (\#1) -- a result ideated from a paper on assessing only accuracy in state prediction for virtual environments \cite{wang-etal-2024-language}, and whose benchmark was unavailable in the sandbox, causing the experiment builder to crawl one of the environments that was accessible to it with a random agent to create \textit{its own benchmark} for the experiment.  Similarly (and also on state-prediction), a different discovery suggested that an LLM's performance on state prediction tasks varies with the complexity of the representation it has to predict (\#2).  
Another discovery found that language models are particularly poor at assessing whether an action will be successful in an environment given the previous environmental observation (\#5).  While language models' arithmetic performance is well studied \citep[e.g.][]{Yuan2023HowWD}, a discovery suggested they are poor at solving a specific combinatorial optimization problem involving addition (\#4). Reports and code for these discoveries are provided in \textsc{Appendix}~\ref{sec:experiment-reports}, with assessments of (incremental) novelty in \textsc{Appendix}~\ref{sec:novelty-assessment}.

{\flushleft\textbf{Rejected discoveries:}} 13 of 19 discoveries were rejected after human evaluation.  These include 6 reported discoveries where the baselines or models themselves were overly simplistic, and 3 reported discoveries where the reported effect disappeared after rerunning the experiment with a larger number of samples.  2 reported discoveries had major implementation errors discovered by the domain expert, and were rejected.  One candidate discovery was rejected by external reviewers for having limitd details from which to evaluate its soundness.

\begin{table}[t!]
\centering
\footnotesize
    \begin{tabular}{lc}
    \toprule
    \textbf{Experiment Builder Outcomes}    &   \%   \\
    \midrule
    ~~Experiment completed successfully       &   41\%  \\    
    ~~Debug iteration limit reached           &   32\%  \\
    ~~Hard experiment time limit reached      &   18\%  \\
    ~~Unrecoverable code generation issue     &   9\%  \\ 
    ~~~~\textit{(i.e. code too long for output)}\\
    ~~Hard cost limit reached                 &   0\%  \\
    \midrule
    Number of samples (experiments)           & 250 \\
    \bottomrule
    \end{tabular}
    \caption{Summary statistics of the \textit{experiment builder}. \label{tab:experiment-builder-statistics}}    
    \vspace{-2mm}
\end{table}

\section{Discussion}
\label{sec:failures-and-challenges}

We outline challenges and common failure cases below, to highlight where targeted efforts might improve discovery quality. %

{\flushleft\textbf{Idea Diversity:}} While the candidate discoveries demonstrate that the ideator can generate diverse ideas that span designing new agents, tasks, metrics, methods, and benchmarks, most of the ideas that are generated are highly similar, or mechanical variations of one another (e.g. \textit{apply method X to benchmark Y}).  Si et al.~\shortcite{si2024can} observed that LLM-ideators quickly saturate the number of unique ideas they generate. We found that using their method of uniqueness filtering (using cosine similarity over embeddings) still generated many duplicates, and had to use manual filtering to select a diverse, fairly non-overlapping subset to explore.

{\flushleft\textbf{Experiment Builder Failures:}} 
Currently, 59\% of the experiments the planner designs are not able to be successfully implemented by the experiment builder.  The distribution of executor errors is shown in Table~\ref{tab:experiment-builder-statistics}. %
The majority of experiments ending in failure face challenges in debugging that are never resolved, hitting either the maximum number of debug iterations (32\%) or experiment time limits (18\%).  Infrequently, the experiment code is too long to fully generate (9\%), requiring a model that can generate more than \textsc{8k} output tokens.

{\flushleft\textbf{Unfaithful Experiments:}} 
The system occasionally reports a result that (upon closer examination of the code) is untrue. For example, the system reported that a graph-based agent with a verification mechanism outperformed a baseline agent \textit{(Result \#12 in Table~\ref{tab:discoveries})}. Human inspection of the code revealed that while the agent built the graph, it never used it, instead picking actions randomly.  These types of errors are difficult and laborious to detect, requiring a great deal of human effort.  Methods to automate this will help reduce false positives/negatives.

{\flushleft\textbf{Adherence to best research practices:}}
Many generated experiments do not adhere to best practices in research methods, even when prompted to do so.  Experiments may evaluate on the training set, or make errors in calculating or interpreting statistics. They may similarly choose weak baselines to compare experimental models against (e.g. comparing modified \textsc{ReAct} agents to random baselines rather than \textsc{ReAct} baselines, as in Table~\ref{tab:discoveries}), erroneously producing statistically significant results.\footnote{In our detailed code analysis, we overlooked research methods issues that were unlikely to invalidate results -- for example, evaluating on the training set with \textit{zero-shot} methods that do not fit any parameters using the training set -- though this is unconventional, and limits the ability to compare these results to those reported in other papers.}

{\flushleft\textbf{Reducing Human Effort:}} Our system currently uses human involvement at 5 steps -- a pragmatic requirement given that when run in fully-automated mode, an early pilot of \codesci generated results much less efficiently: ideas were frequently duplicated, experiments failed more often during debugging, and results had methods challenges and were less convincing.  Currently human involvement is included in selecting papers to ideate from, generating codeblocks, selecting interesting ideas to run, providing brief expert comments on each ideas, and performing a manual validation of the results. %
Towards the goal of automation, a list of representative papers could be provided by a paper selection agent in response to a broad area of interest provided by a user.  Many of the codeblocks used by the system were already produced by a language model, but manually corrected and tweaked by a human for running in the sandbox.  The rapid progress in ideation models~\citep[e.g.][]{Wang2023SciMONSI} and evaluation \cite{Radensky2024ScideatorHS} is likely to produce systems that continue to increase idea diversity and utility in the near-term, reducing the need for a human to filter these ideas.  Our use of brief domain-expert comments on ideas before generating plans helps the system shore up obvious research methods issues (like using a better metric) before running costly experiments, and as models increase their knowledge of domain-specific research methods, this reliance will decrease. %

{\flushleft\textbf{Fully-automated Mode:}} While we present \codesci as requiring human input, it is possible to run \codesci in a fully-automated mode without such input.  Here we present two ablations that remove human input, with the caveats that (a) the variance of the system (even when provided with identical experiment plans) is high, (b) the overall candidate discovery rate is low, and (c) large-scale runs are prohibitively expensive -- which, taken together, makes performing ablations with enough statistical power to make inferences currently impractical.  That being said, in pilot experiments, running \codesci on 100 ideas generated and executed autonomously (without human input) appeared to generate 2 candidate discoveries that passed an internal review, or a 2\% success rate (and, our motivation for pre-filtering ideas to those most likely to produce discoveries).  Similarly, rerunning the benchmark of 50 ideas described in Section~\ref{sec:discovery-experiments} \textit{without} expert human comments appears to produce 4 of the same 6 candidate discoveries as were found in Table~\ref{tab:discoveries} -- nominally reducing the number of candidate discoveries by approximately a third.  It is important to note that this is a small sample, and difficult to ascertain whether this reduction is due to the lack of expert comments, or the natural variance in the system.

\section{Conclusion}
We present \codesci, an end-to-end semi-automated discovery system with a novel ideator that performs genetic search jointly on combinations of literature and codeblocks, before building, running, and analyzing these software artifacts in experiments.  Ideating from literature and common codeblocks in the domain of \textit{agents and environments}, \codesci identified 19 potential discoveries, 6 of which were judged as meeting minimum thresholds for scientific soundness and incremental novelty after both external conference-style review, and an internal code review.  A qualitative review of the discoveries suggests that they broaden the scope of ideas explored by other systems, with candidate discoveries spanning novel tasks, agents, metrics, and data.  We provide an analysis of common failure modes, and release our system as open-source to facilitate future research in automated scientific discovery.

\section{Limitations}

We highlight the following limitations both in this work, as well as in current automated discovery systems more broadly. 

{\flushleft\textbf{Cost vs accuracy trade-off:}}
The automated experiments in this work are designed to be fast, inexpensive estimates of an experiment's results, that allow rapid iteration. The average experiment in this work costs $\approx$\$4, and takes approximately 2 hours to complete. While these experiments save on resources, due to their low number of samples, they both produce false positives, and are likely unable to detect all but the largest effects (producing false negatives). This is not a technical limitation of this work as (in principle) it can be remedied by scaling the computation budget -- though developing strategies for intelligencly investing in the experiments most likely to have utility could reduce budget requirements.

{\flushleft\textbf{Validating Candidate Discoveries:}}
The common mode of disseminating research artifacts in \textsc{NLP/AI} is through peer-reviewed scientific articles.  This peer review typically examines the article in detail, but reviewers are frequently instructed that they are not required to provide a review of supplementary material including code\footnote{\url{https://aclrollingreview.org/reviewerguidelines}} -- likely in part because code review is extremely labor intensive, but also because a degree of skill, domain-specific training, and a good-faith effort are assumed on behalf of the authors.  The opposite appears true of language-model-generated code, and in this work we show that more than half of the potential discoveries were rejected by an internal code review by a domain expert (one of the authors) for having serious issues.  We observed -- both in pilot experiments, as well as in evaluating the 19 potential discoveries -- that the LLM-generated code may unfaithfully represent the desired process or mechanism requested in the experiment plan, and instead actually perform a much simpler procedure (like randomly generating actions to take in a virtual environment). These are particularly challenging to find because, often, much of what the LLM claims is happening is at least partially implemented, and the offending code may be only one (or a few) lines in a long program.  As such, while we have made a good-faith effort to validate the discoveries and code proposed by \codesci -- including examining the code and rerunning each discovery at a greater scale, which is arguably a more effortful review than is done for most peer-reviewed articles -- given the sometimes adversarial nature of the task, it is possible we may not have discovered some inaccurately reported mechanisms.  As such, we frame the 6 discoveries made by the system (and having passed external and internal review) as ``candidate discoveries'', to emphasize their preliminary nature, and the challenges of performing review on long LLM-generated code.  Developing automated mechanisms to speed this review is likely of paramount importance to scaling automated scientific discovery systems that use code-based experimentation.

{\flushleft\textbf{Incremental vs Transformational Discoveries:}} 
The 6 expert-validated discoveries produced by \codesci would likely be categorized by most as normal incremental science rather than transformational discoveries -- and the ratings by the 3 external reviewers suggest that each candidate \codesci discovery is (at best) incrementally novel.  For example, while the first result (\#1) in Table~\ref{tab:discoveries} finds that an LLM's self-assessed confidence in its predictions has a low correlation with its actual performance on state prediction tasks focusing on using LLMs as world models, similar effects have been shown on other tasks, and testing the existence of this phenomenon on state prediction in virtual environments is an incremental discovery.  The discoveries produced by \codesci are incremental rather than transformational discoveries, and we have no data to support whether generating more impactful discoveries is a problem of scale (i.e. running either more ideas, or higher risk/higher gain ideas), or a problem of kind (i.e. whether ideation and execution in the manner that we have described here is capable or incapable of generating high-impact discoveries).  

{\flushleft\textbf{Ideator Recall:}} Studying the ``recall'' of an ideator (in terms of the proportion of ideas that it generates that ultimately provide discoveries that are judged to be at least minimally sound and incrementally novel) is an open area of research.  In pilot experiments we generated new ideas at runtime, many of which had issues (such as being near duplicates, using incorrect metrics, or being very challenging to implement).  As a pragmatic cost saving measure, we explicitly filter down a large set of ideas to manually select the first 50 ideas that appeared to meet the bar of being potentially implementable, relatively different from one another, and that did not appear to have obvious research methods problems. As such, due to this cost saving measure, we are unable to make claims about what proportion of ideas that the ideator generates ultimately lead to human-verified discoveries.

\bibliography{anthology,custom}

\begin{thebibliography}{33}
\providecommand{\natexlab}[1]{#1}

\bibitem[{Chen et~al.(2023)Chen, Liang, Huang, Real, Wang, Liu, Pham, Dong, Luong, Hsieh, Lu, and Le}]{lion}
Xiangning Chen, Chen Liang, Da~Huang, Esteban Real, Kaiyuan Wang, Yao Liu, Hieu Pham, Xuanyi Dong, Thang Luong, Cho-Jui Hsieh, Yifeng Lu, and Quoc~V. Le. 2023.
\newblock Symbolic discovery of optimization algorithms.
\newblock \emph{ArXiv}, abs/2302.06675.

\bibitem[{Efron(1992)}]{efron1992bootstrap}
Bradley Efron. 1992.
\newblock Bootstrap methods: another look at the jackknife.
\newblock In \emph{Breakthroughs in statistics: Methodology and distribution}, pages 569--593. Springer.

\bibitem[{Geng et~al.(2024)Geng, Cai, Wang, Koeppl, Nakov, and Gurevych}]{geng-etal-2024-survey}
Jiahui Geng, Fengyu Cai, Yuxia Wang, Heinz Koeppl, Preslav Nakov, and Iryna Gurevych. 2024.
\newblock \href {https://doi.org/10.18653/v1/2024.naacl-long.366} {A survey of confidence estimation and calibration in large language models}.
\newblock In \emph{Proceedings of the 2024 Conference of the North American Chapter of the Association for Computational Linguistics: Human Language Technologies (Volume 1: Long Papers)}, pages 6577--6595, Mexico City, Mexico. Association for Computational Linguistics.

\bibitem[{Gu et~al.(2023)Gu, Dalvi~Mishra, and Clark}]{gu-etal-2023-language}
Yuling Gu, Bhavana Dalvi~Mishra, and Peter Clark. 2023.
\newblock \href {https://doi.org/10.18653/v1/2023.acl-long.106} {Do language models have coherent mental models of everyday things?}
\newblock In \emph{Proceedings of the 61st Annual Meeting of the Association for Computational Linguistics (Volume 1: Long Papers)}, pages 1892--1913, Toronto, Canada. Association for Computational Linguistics.

\bibitem[{Hemberg et~al.(2024)Hemberg, Moskal, and O’Reilly}]{llm-gp}
Erik Hemberg, Stephen Moskal, and Una-May O’Reilly. 2024.
\newblock Evolving code with a large language model.
\newblock \emph{Genet. Program. Evolvable Mach.}, 25:21.

\bibitem[{Huang et~al.(2024)Huang, Vora, Liang, and Leskovec}]{mlagentbench}
Qian Huang, Jian Vora, Percy Liang, and Jure Leskovec. 2024.
\newblock Mlagentbench: Evaluating language agents on machine learning experimentation.
\newblock In \emph{Forty-first International Conference on Machine Learning}.

\bibitem[{Hunter(2007)}]{hunter2007matplotlib}
John~D Hunter. 2007.
\newblock Matplotlib: A 2d graphics environment.
\newblock \emph{Computing in science \& engineering}, 9(03):90--95.

\bibitem[{Ifargan et~al.(2024)Ifargan, Hafner, Kern, Alcalay, and Kishony}]{DataToPaper}
Tal Ifargan, Lukas Hafner, Maor Kern, Ori Alcalay, and Roy Kishony. 2024.
\newblock Autonomous llm-driven research from data to human-verifiable research papers.
\newblock \emph{ArXiv}, abs/2404.17605.

\bibitem[{{Intology AI}(2025)}]{zochi}
{Intology AI}. 2025.
\newblock \href {{https://github.com/IntologyAI/Zochi/blob/main/Zochi_Technical_Report.pdf}} {Zochi technical report}.
\newblock Technical report, {Intology AI}.

\bibitem[{Jansen(2022)}]{jansen-2022-systematic}
Peter Jansen. 2022.
\newblock \href {https://doi.org/10.18653/v1/2022.wordplay-1.1} {A systematic survey of text worlds as embodied natural language environments}.
\newblock In \emph{Proceedings of the 3rd Wordplay: When Language Meets Games Workshop (Wordplay 2022)}, pages 1--15, Seattle, United States. Association for Computational Linguistics.

\bibitem[{Jansen and Cote(2023)}]{jansen-cote-2023-textworldexpress}
Peter Jansen and Marc-alexandre Cote. 2023.
\newblock \href {https://doi.org/10.18653/v1/2023.eacl-demo.20} {{T}ext{W}orld{E}xpress: Simulating text games at one million steps per second}.
\newblock In \emph{Proceedings of the 17th Conference of the European Chapter of the Association for Computational Linguistics: System Demonstrations}, pages 169--177, Dubrovnik, Croatia. Association for Computational Linguistics.

\bibitem[{Jansen et~al.(2024)Jansen, C{\^o}t{\'e}, Khot, Bransom, Mishra, Majumder, Tafjord, and Clark}]{discoveryworld}
Peter Jansen, Marc-Alexandre C{\^o}t{\'e}, Tushar Khot, Erin Bransom, Bhavana~Dalvi Mishra, Bodhisattwa~Prasad Majumder, Oyvind Tafjord, and Peter Clark. 2024.
\newblock \href {https://openreview.net/forum?id=cDYqckEt6d} {Discoveryworld: A virtual environment for developing and evaluating automated scientific discovery agents}.
\newblock In \emph{The Thirty-eight Conference on Neural Information Processing Systems Datasets and Benchmarks Track}.

\bibitem[{Jimenez et~al.(2023)Jimenez, Yang, Wettig, Yao, Pei, Press, and Narasimhan}]{swebench}
Carlos~E Jimenez, John Yang, Alexander Wettig, Shunyu Yao, Kexin Pei, Ofir Press, and Karthik~R Narasimhan. 2023.
\newblock Swe-bench: Can language models resolve real-world github issues?
\newblock In \emph{The Twelfth International Conference on Learning Representations}.

\bibitem[{Jumper et~al.(2021)Jumper, Evans, Pritzel, Green, Figurnov, Ronneberger, Tunyasuvunakool, Bates, Ž{\'i}dek, Potapenko, Bridgland, Meyer, Kohl, Ballard, Cowie, Romera-Paredes, Nikolov, Jain, Adler, Back, Petersen, Reiman, Clancy, Zielinski, Steinegger, Pacholska, Berghammer, Bodenstein, Silver, Vinyals, Senior, Kavukcuoglu, Kohli, and Hassabis}]{alphafold}
John~M. Jumper, Richard Evans, Alexander Pritzel, Tim Green, Michael Figurnov, Olaf Ronneberger, Kathryn Tunyasuvunakool, Russ Bates, Augustin Ž{\'i}dek, Anna Potapenko, Alex Bridgland, Clemens Meyer, Simon A~A Kohl, Andy Ballard, Andrew Cowie, Bernardino Romera-Paredes, Stanislav Nikolov, Rishub Jain, Jonas Adler, Trevor Back, Stig Petersen, David Reiman, Ellen Clancy, Michal Zielinski, Martin Steinegger, Michalina Pacholska, Tamas Berghammer, Sebastian Bodenstein, David Silver, Oriol Vinyals, Andrew~W. Senior, Koray Kavukcuoglu, Pushmeet Kohli, and Demis Hassabis. 2021.
\newblock Highly accurate protein structure prediction with alphafold.
\newblock \emph{Nature}, 596:583 -- 589.

\bibitem[{Li et~al.(2024)Li, Patel, Wang, Wang, and Du}]{Li2024MLRCopilotAM}
Ruochen Li, Teerth Patel, Qingyun Wang, Qingyun Wang, and Xinya Du. 2024.
\newblock \href {https://api.semanticscholar.org/CorpusID:271957477} {Mlr-copilot: Autonomous machine learning research based on large language models agents}.
\newblock \emph{ArXiv}, abs/2408.14033.

\bibitem[{Likert(1932)}]{likert1932technique}
Rensis Likert. 1932.
\newblock A technique for the measurement of attitudes.
\newblock \emph{Archives of Psychology}.

\bibitem[{Liu et~al.(2024)Liu, Liu, Zhu, Lei, Yang, Zhang, Li, and Liu}]{aigs}
Zijun Liu, Kai Liu, Yiqi Zhu, Xuanyu Lei, Zonghan Yang, Zhenhe Zhang, Peng Li, and Yang Liu. 2024.
\newblock Aigs: Generating science from ai-powered automated falsification.
\newblock \emph{ArXiv}, abs/2411.11910.

\bibitem[{Lu et~al.(2024{\natexlab{a}})Lu, Lu, Lange, Foerster, Clune, and Ha}]{aiscientist}
Chris Lu, Cong Lu, Robert~Tjarko Lange, Jakob~N. Foerster, Jeff Clune, and David Ha. 2024{\natexlab{a}}.
\newblock The ai scientist: Towards fully automated open-ended scientific discovery.
\newblock \emph{ArXiv}, abs/2408.06292.

\bibitem[{Lu et~al.(2024{\natexlab{b}})Lu, Lu, Lange, Foerster, Clune, and Ha}]{Lu2024TheAS}
Chris Lu, Cong Lu, Robert~Tjarko Lange, Jakob~N. Foerster, Jeff Clune, and David Ha. 2024{\natexlab{b}}.
\newblock \href {https://api.semanticscholar.org/CorpusID:271854887} {The ai scientist: Towards fully automated open-ended scientific discovery}.
\newblock \emph{ArXiv}, abs/2408.06292.

\bibitem[{Madaan et~al.(2023)Madaan, Tandon, Gupta, Hallinan, Gao, Wiegreffe, Alon, Dziri, Prabhumoye, Yang et~al.}]{madaanself}
Aman Madaan, Niket Tandon, Prakhar Gupta, Skyler Hallinan, Luyu Gao, Sarah Wiegreffe, Uri Alon, Nouha Dziri, Shrimai Prabhumoye, Yiming Yang, et~al. 2023.
\newblock Self-refine: Iterative refinement with self-feedback.
\newblock In \emph{Thirty-seventh Conference on Neural Information Processing Systems}.

\bibitem[{Miller(1995)}]{miller1995wordnet}
George~A Miller. 1995.
\newblock Wordnet: a lexical database for english.
\newblock \emph{Communications of the ACM}, 38(11):39--41.

\bibitem[{OpenAI et~al.(2024)OpenAI, Achiam, Adler, Agarwal, Ahmad, Akkaya, Aleman, Almeida, Altenschmidt, Altman, Anadkat, Avila, Babuschkin, Balaji, Balcom, Baltescu, Bao, Bavarian, Belgum, Bello, Berdine, Bernadett-Shapiro, Berner, Bogdonoff, Boiko, Boyd, Brakman, Brockman, Brooks, Brundage, Button, Cai, Campbell, Cann, Carey, Carlson, Carmichael, Chan, Chang, Chantzis, Chen, Chen, Chen, Chen, Chen, Chess, Cho, Chu, Chung, Cummings, Currier, Dai, Decareaux, Degry, Deutsch, Deville, Dhar, Dohan, Dowling, Dunning, Ecoffet, Eleti, Eloundou, Farhi, Fedus, Felix, Fishman, Forte, Fulford, Gao, Georges, Gibson, Goel, Gogineni, Goh, Gontijo-Lopes, Gordon, Grafstein, Gray, Greene, Gross, Gu, Guo, Hallacy, Han, Harris, He, Heaton, Heidecke, Hesse, Hickey, Hickey, Hoeschele, Houghton, Hsu, Hu, Hu, Huizinga, Jain, Jain, Jang, Jiang, Jiang, Jin, Jin, Jomoto, Jonn, Jun, Kaftan, Łukasz Kaiser, Kamali, Kanitscheider, Keskar, Khan, Kilpatrick, Kim, Kim, Kim, Kirchner, Kiros, Knight, Kokotajlo, Łukasz Kondraciuk,
  Kondrich, Konstantinidis, Kosic, Krueger, Kuo, Lampe, Lan, Lee, Leike, Leung, Levy, Li, Lim, Lin, Lin, Litwin, Lopez, Lowe, Lue, Makanju, Malfacini, Manning, Markov, Markovski, Martin, Mayer, Mayne, McGrew, McKinney, McLeavey, McMillan, McNeil, Medina, Mehta, Menick, Metz, Mishchenko, Mishkin, Monaco, Morikawa, Mossing, Mu, Murati, Murk, Mély, Nair, Nakano, Nayak, Neelakantan, Ngo, Noh, Ouyang, O'Keefe, Pachocki, Paino, Palermo, Pantuliano, Parascandolo, Parish, Parparita, Passos, Pavlov, Peng, Perelman, de~Avila Belbute~Peres, Petrov, de~Oliveira~Pinto, Michael, Pokorny, Pokrass, Pong, Powell, Power, Power, Proehl, Puri, Radford, Rae, Ramesh, Raymond, Real, Rimbach, Ross, Rotsted, Roussez, Ryder, Saltarelli, Sanders, Santurkar, Sastry, Schmidt, Schnurr, Schulman, Selsam, Sheppard, Sherbakov, Shieh, Shoker, Shyam, Sidor, Sigler, Simens, Sitkin, Slama, Sohl, Sokolowsky, Song, Staudacher, Such, Summers, Sutskever, Tang, Tezak, Thompson, Tillet, Tootoonchian, Tseng, Tuggle, Turley, Tworek, Uribe, Vallone,
  Vijayvergiya, Voss, Wainwright, Wang, Wang, Wang, Ward, Wei, Weinmann, Welihinda, Welinder, Weng, Weng, Wiethoff, Willner, Winter, Wolrich, Wong, Workman, Wu, Wu, Wu, Xiao, Xu, Yoo, Yu, Yuan, Zaremba, Zellers, Zhang, Zhang, Zhao, Zheng, Zhuang, Zhuk, and Zoph}]{openai2024gpt4technicalreport}
OpenAI, Josh Achiam, Steven Adler, Sandhini Agarwal, Lama Ahmad, Ilge Akkaya, Florencia~Leoni Aleman, Diogo Almeida, Janko Altenschmidt, Sam Altman, Shyamal Anadkat, Red Avila, Igor Babuschkin, Suchir Balaji, Valerie Balcom, Paul Baltescu, Haiming Bao, Mohammad Bavarian, Jeff Belgum, Irwan Bello, Jake Berdine, Gabriel Bernadett-Shapiro, Christopher Berner, Lenny Bogdonoff, Oleg Boiko, Madelaine Boyd, Anna-Luisa Brakman, Greg Brockman, Tim Brooks, Miles Brundage, Kevin Button, Trevor Cai, Rosie Campbell, Andrew Cann, Brittany Carey, Chelsea Carlson, Rory Carmichael, Brooke Chan, Che Chang, Fotis Chantzis, Derek Chen, Sully Chen, Ruby Chen, Jason Chen, Mark Chen, Ben Chess, Chester Cho, Casey Chu, Hyung~Won Chung, Dave Cummings, Jeremiah Currier, Yunxing Dai, Cory Decareaux, Thomas Degry, Noah Deutsch, Damien Deville, Arka Dhar, David Dohan, Steve Dowling, Sheila Dunning, Adrien Ecoffet, Atty Eleti, Tyna Eloundou, David Farhi, Liam Fedus, Niko Felix, Simón~Posada Fishman, Juston Forte, Isabella Fulford, Leo
  Gao, Elie Georges, Christian Gibson, Vik Goel, Tarun Gogineni, Gabriel Goh, Rapha Gontijo-Lopes, Jonathan Gordon, Morgan Grafstein, Scott Gray, Ryan Greene, Joshua Gross, Shixiang~Shane Gu, Yufei Guo, Chris Hallacy, Jesse Han, Jeff Harris, Yuchen He, Mike Heaton, Johannes Heidecke, Chris Hesse, Alan Hickey, Wade Hickey, Peter Hoeschele, Brandon Houghton, Kenny Hsu, Shengli Hu, Xin Hu, Joost Huizinga, Shantanu Jain, Shawn Jain, Joanne Jang, Angela Jiang, Roger Jiang, Haozhun Jin, Denny Jin, Shino Jomoto, Billie Jonn, Heewoo Jun, Tomer Kaftan, Łukasz Kaiser, Ali Kamali, Ingmar Kanitscheider, Nitish~Shirish Keskar, Tabarak Khan, Logan Kilpatrick, Jong~Wook Kim, Christina Kim, Yongjik Kim, Jan~Hendrik Kirchner, Jamie Kiros, Matt Knight, Daniel Kokotajlo, Łukasz Kondraciuk, Andrew Kondrich, Aris Konstantinidis, Kyle Kosic, Gretchen Krueger, Vishal Kuo, Michael Lampe, Ikai Lan, Teddy Lee, Jan Leike, Jade Leung, Daniel Levy, Chak~Ming Li, Rachel Lim, Molly Lin, Stephanie Lin, Mateusz Litwin, Theresa Lopez, Ryan
  Lowe, Patricia Lue, Anna Makanju, Kim Malfacini, Sam Manning, Todor Markov, Yaniv Markovski, Bianca Martin, Katie Mayer, Andrew Mayne, Bob McGrew, Scott~Mayer McKinney, Christine McLeavey, Paul McMillan, Jake McNeil, David Medina, Aalok Mehta, Jacob Menick, Luke Metz, Andrey Mishchenko, Pamela Mishkin, Vinnie Monaco, Evan Morikawa, Daniel Mossing, Tong Mu, Mira Murati, Oleg Murk, David Mély, Ashvin Nair, Reiichiro Nakano, Rajeev Nayak, Arvind Neelakantan, Richard Ngo, Hyeonwoo Noh, Long Ouyang, Cullen O'Keefe, Jakub Pachocki, Alex Paino, Joe Palermo, Ashley Pantuliano, Giambattista Parascandolo, Joel Parish, Emy Parparita, Alex Passos, Mikhail Pavlov, Andrew Peng, Adam Perelman, Filipe de~Avila Belbute~Peres, Michael Petrov, Henrique~Ponde de~Oliveira~Pinto, Michael, Pokorny, Michelle Pokrass, Vitchyr~H. Pong, Tolly Powell, Alethea Power, Boris Power, Elizabeth Proehl, Raul Puri, Alec Radford, Jack Rae, Aditya Ramesh, Cameron Raymond, Francis Real, Kendra Rimbach, Carl Ross, Bob Rotsted, Henri Roussez,
  Nick Ryder, Mario Saltarelli, Ted Sanders, Shibani Santurkar, Girish Sastry, Heather Schmidt, David Schnurr, John Schulman, Daniel Selsam, Kyla Sheppard, Toki Sherbakov, Jessica Shieh, Sarah Shoker, Pranav Shyam, Szymon Sidor, Eric Sigler, Maddie Simens, Jordan Sitkin, Katarina Slama, Ian Sohl, Benjamin Sokolowsky, Yang Song, Natalie Staudacher, Felipe~Petroski Such, Natalie Summers, Ilya Sutskever, Jie Tang, Nikolas Tezak, Madeleine~B. Thompson, Phil Tillet, Amin Tootoonchian, Elizabeth Tseng, Preston Tuggle, Nick Turley, Jerry Tworek, Juan Felipe~Cerón Uribe, Andrea Vallone, Arun Vijayvergiya, Chelsea Voss, Carroll Wainwright, Justin~Jay Wang, Alvin Wang, Ben Wang, Jonathan Ward, Jason Wei, CJ~Weinmann, Akila Welihinda, Peter Welinder, Jiayi Weng, Lilian Weng, Matt Wiethoff, Dave Willner, Clemens Winter, Samuel Wolrich, Hannah Wong, Lauren Workman, Sherwin Wu, Jeff Wu, Michael Wu, Kai Xiao, Tao Xu, Sarah Yoo, Kevin Yu, Qiming Yuan, Wojciech Zaremba, Rowan Zellers, Chong Zhang, Marvin Zhang, Shengjia
  Zhao, Tianhao Zheng, Juntang Zhuang, William Zhuk, and Barret Zoph. 2024.
\newblock \href {https://arxiv.org/abs/2303.08774} {Gpt-4 technical report}.
\newblock \emph{Preprint}, arXiv:2303.08774.

\bibitem[{Qi et~al.(2024)Qi, Jia, Liu, Zhan, Zhang, Wen, Gan, Chen, Liu, Ma, Li, Wang, Kulkarni, Chen, Zhou, Li, Wang, and Huang}]{Qi2024MetaScientistAH}
Jingyuan Qi, Zian Jia, Minqian Liu, Wangzhi Zhan, Junkai Zhang, Xiaofei Wen, Jingru Gan, Jianpeng Chen, Qin Liu, Mingyu~Derek Ma, Bangzheng Li, Haohui Wang, Adithya Kulkarni, Muhao Chen, Dawei Zhou, Ling Li, Wei Wang, and Lifu Huang. 2024.
\newblock \href {https://api.semanticscholar.org/CorpusID:274981805} {Metascientist: A human-ai synergistic framework for automated mechanical metamaterial design}.
\newblock \emph{ArXiv}, abs/2412.16270.

\bibitem[{Radensky et~al.(2024)Radensky, Shahid, Fok, Siangliulue, Hope, and Weld}]{Radensky2024ScideatorHS}
Marissa Radensky, Simra Shahid, Raymond Fok, Pao Siangliulue, Tom Hope, and Daniel~S. Weld. 2024.
\newblock \href {https://api.semanticscholar.org/CorpusID:272827497} {Scideator: Human-llm scientific idea generation grounded in research-paper facet recombination}.
\newblock \emph{ArXiv}, abs/2409.14634.

\bibitem[{Schmidgall et~al.(2025)Schmidgall, Su, Wang, Sun, Wu, Yu, Liu, Liu, and Barsoum}]{agentlab}
Samuel Schmidgall, Yusheng Su, Ze~Wang, Ximeng Sun, Jialian Wu, Xiaodong Yu, Jiang Liu, Zicheng Liu, and Emad Barsoum. 2025.
\newblock \href {https://api.semanticscholar.org/CorpusID:275358017} {Agent laboratory: Using llm agents as research assistants}.
\newblock In \emph{arXiv}, volume abs/2501.04227.

\bibitem[{Si et~al.(2024)Si, Yang, and Hashimoto}]{si2024can}
Chenglei Si, Diyi Yang, and Tatsunori Hashimoto. 2024.
\newblock Can llms generate novel research ideas? a large-scale human study with 100+ nlp researchers.
\newblock \emph{arXiv preprint arXiv:2409.04109}.

\bibitem[{Speer et~al.(2017)Speer, Chin, and Havasi}]{speer2017conceptnet}
Robyn Speer, Joshua Chin, and Catherine Havasi. 2017.
\newblock Conceptnet 5.5: An open multilingual graph of general knowledge.
\newblock In \emph{Proceedings of the AAAI conference on artificial intelligence}, volume~31.

\bibitem[{Stokes et~al.(2020)Stokes, Yang, Swanson, Jin, Cubillos-Ruiz, Donghia, MacNair, French, Carfrae, Bloom-Ackermann, Tran, Chiappino-Pepe, Badran, Andrews, Chory, Church, Brown, Jaakkola, Barzilay, and Collins}]{halicin}
Jonathan~M. Stokes, Kevin Yang, Kyle Swanson, Wengong Jin, Andr{\'e}s Cubillos-Ruiz, Nina~M. Donghia, Craig~R. MacNair, Shawn French, Lindsey~A. Carfrae, Zohar Bloom-Ackermann, Victoria~M. Tran, Anush Chiappino-Pepe, Ahmed~H. Badran, Ian~W. Andrews, Emma~J. Chory, George~M. Church, Eric~D. Brown, T.~Jaakkola, Regina Barzilay, and James~J. Collins. 2020.
\newblock A deep learning approach to antibiotic discovery.
\newblock \emph{Cell}, 181:475--483.

\bibitem[{Wang et~al.(2023{\natexlab{a}})Wang, Downey, Ji, and Hope}]{Wang2023SciMONSI}
Qingyun Wang, Doug Downey, Heng Ji, and Tom Hope. 2023{\natexlab{a}}.
\newblock \href {https://api.semanticscholar.org/CorpusID:258841365} {Scimon: Scientific inspiration machines optimized for novelty}.
\newblock In \emph{Annual Meeting of the Association for Computational Linguistics}.

\bibitem[{Wang et~al.(2024)Wang, Todd, Xiao, Yuan, C{\^o}t{\'e}, Clark, and Jansen}]{wang-etal-2024-language}
Ruoyao Wang, Graham Todd, Ziang Xiao, Xingdi Yuan, Marc-Alexandre C{\^o}t{\'e}, Peter Clark, and Peter Jansen. 2024.
\newblock \href {https://doi.org/10.18653/v1/2024.acl-short.1} {Can language models serve as text-based world simulators?}
\newblock In \emph{Proceedings of the 62nd Annual Meeting of the Association for Computational Linguistics (Volume 2: Short Papers)}, pages 1--17, Bangkok, Thailand. Association for Computational Linguistics.

\bibitem[{Wang et~al.(2023{\natexlab{b}})Wang, Todd, Yuan, Xiao, C{\^o}t{\'e}, and Jansen}]{wang-etal-2023-bytesized32}
Ruoyao Wang, Graham Todd, Xingdi Yuan, Ziang Xiao, Marc-Alexandre C{\^o}t{\'e}, and Peter Jansen. 2023{\natexlab{b}}.
\newblock \href {https://doi.org/10.18653/v1/2023.emnlp-main.830} {{B}yte{S}ized32: A corpus and challenge task for generating task-specific world models expressed as text games}.
\newblock In \emph{Proceedings of the 2023 Conference on Empirical Methods in Natural Language Processing}, pages 13455--13471, Singapore. Association for Computational Linguistics.

\bibitem[{Yao et~al.(2023)Yao, Zhao, Yu, Du, Shafran, Narasimhan, and Cao}]{yaoreact}
Shunyu Yao, Jeffrey Zhao, Dian Yu, Nan Du, Izhak Shafran, Karthik~R Narasimhan, and Yuan Cao. 2023.
\newblock React: Synergizing reasoning and acting in language models.
\newblock In \emph{The Eleventh International Conference on Learning Representations}.

\bibitem[{Yuan et~al.(2023)Yuan, Yuan, Tan, Wang, and Huang}]{Yuan2023HowWD}
Zheng Yuan, Hongyi Yuan, Chuanqi Tan, Wei Wang, and Songfang Huang. 2023.
\newblock \href {https://api.semanticscholar.org/CorpusID:257952500} {How well do large language models perform in arithmetic tasks?}
\newblock \emph{ArXiv}, abs/2304.02015.

\end{thebibliography}

\clearpage
\appendix
\label{sec:appendix}
\section*{Appendix}
\section{Table of Contents}
Below is a list of links to major sections:

\begin{enumerate}
    \item[] \textbf{Additional System Details}            
    \item \hyperref[sec:external-reviewer-rubric]{External Reviewer Rubric}
    \item \hyperref[sec:meta-analysis-categorization-criteria]{Meta-Analysis Classification Criteria}
    \item \hyperref[sec:domain-expert-comments]{All Domain-Expert Comments}    
        \item \hyperref[tab:ideation-operationalization-example]{Example Idea, Comment, \& Plan}    
    \item[] \textbf{Prompts}
    \item \hyperref[sec:prompt-ideation]{Ideation}
    \item \hyperref[sec:appendix-operationalization-prompt]{Planning}
    \item \hyperref[sec:prompt-experiment-builder]{Experiment Builder}
    \item \hyperref[sec:prompt-report-builder]{Reporting}
    \item \hyperref[sec:prompt-summary]{Experiment Summary}
    \item \hyperref[sec:prompt-meta-analysis]{Meta-Analysis}        
    \item[] \textbf{Novelty Assessments of Discoveries}
    \item \hyperref[sec:novelty-assessment]{Novelty Assessments of Discoveries}
    \item[] \textbf{Example Experiment Reports}
    \item \hyperref[report1]{State Prediction Confidence (Report)}    
    \item \hyperref[report2]{Progressive State Complexity (Report)}
    \item \hyperref[report3]{Graph Alignment Metric (Report)}    
    \item \hyperref[report4]{Multi-Stage Environment Generation (Report)}
    \item \hyperref[report5]{Simulation Confidence (Report)}
    \item \hyperref[report6]{Graph Agent for Discovery (Report)}
    \item \hyperref[report7]{Combinatorial Optimization (Report)}    
    \item[] \textbf{Example Experiment Code}    
    \item \hyperref[report1code]{State Prediction Confidence (Code)}
    \item \hyperref[report2code]{Progressive State Complexity (Code)}
    \item \hyperref[report3code]{Graph Alignment Metric (Code)}
    \item \hyperref[report4code]{Multi-Stage Environment Generation (Code)}    
    \item \hyperref[report5code]{Simulation Confidence (Code)}    
    \item \hyperref[report6code]{Graph Agent for Discovery (Code)}    
    \item \hyperref[report7code]{Combinatorial Optimization (Code)}
\end{enumerate}

\section{External Reviewer Rubric}
\label{sec:external-reviewer-rubric}

External reviewers were provided with each of the papers from Table~\ref{tab:discoveries}, and asked to categorize them on 2 categorical scales: scientific soundness and novelty.  In addition, they were asked to provide short justification for their ratings, as well as an overall brief description of the contributions and claims of the paper.

\begin{table}[H]
    \footnotesize
    \centering
    \begin{tabular}{p{0.90\linewidth}}
    \textbf{Soundness} refers to the rigor and reliability of a study’s methods and evidence—essentially, how well the experiments and analysis support the claims made. Please provide a rating for the \textbf{soundness} of this study: \vspace{0.5em}\\
    \textbf{(A) Clearly Sound:} The study demonstrates robust methodology; its design, implementation, and analysis fully support the claims. \vspace{0.5em}\\
    \textbf{(B) Likely Sound:} Assuming faithful execution, the methodology appears sound, with evidence generally supporting the claims despite minor uncertainties. \vspace{0.5em}\\
    \textbf{(C) Minor Concerns:} Identified methodological limitations may slightly affect measurements (e.g., effect sizes) but do not alter the overall conclusions. \vspace{0.5em}\\
    \textbf{(D): Unsound:} Evident methodological or conceptual flaws undermine the credibility of the claims and contributions. \vspace{0.5em}\\
    \end{tabular}
\end{table}

\begin{table}[H]
    \footnotesize
    \centering
    \begin{tabular}{p{0.90\linewidth}}
    \textbf{Novelty} refers to how original or innovative a study’s contributions are relative to existing work. It assesses whether the contributions offer incremental changes or significant departures from what has been done before. Please provide a rating for the \textbf{novelty} of this study. \vspace{0.5em}\\
    \textbf{(A) Highly Novel:} Introduces entirely new concepts or frameworks not previously explored. Example: A modeling contribution that proposes a novel architecture that redefines established paradigms in NLP. \vspace{0.5em}\\
    \textbf{(B) Incrementally Novel (Significant Variation):} Substantially modifies existing approaches, leading to marked advancements. Example: A modeling contribution that makes significant architectural or algorithmic changes that enhance performance. \vspace{0.5em}\\
    \textbf{(C) Incrementally Novel (Minor Variation):} Presents modest modifications or adaptations to established work. Example: A modeling contribution that applies an existing model to a new task with only minor tweaks or parameter adjustments. \vspace{0.5em}\\
    \textbf{(D): Not Novel/Exists in Exact Form:} Replicates existing work without introducing any modifications. Example: A modeling contribution that runs an existing model on an existing task, where the result is already known. \vspace{0.5em}\\
    \end{tabular}
\end{table}

\section{Meta-Analysis Categorization Criteria}
\label{sec:meta-analysis-categorization-criteria}

The following criteria were used to classify a suite of 5 experiments as having either \textbf{consistent}, \textbf{mixed}, or \textbf{limited} results: 

\begin{enumerate}
\item \textbf{Consistent (C):} If at least 80\% (i.e. 4 of the 5) independent experiment runs for a given idea generated the same high-level result (i.e. supporting or rejecting the hypothesis), then it was classified as consistent. 
\item \textbf{Mixed (M):} If a set of 5 experiments was neither classified as \textit{consistent} or \textit{limited}, then it was classified as having \textit{mixed} results. 
\item \textbf{Limited (L):} If 40\% or fewer (i.e. 2 or fewer of the 5 runs) successfully completed, regardless of the outcome of those experiments (i.e. support, reject, or inconclusive towards the hypothesis), then it was classified as limited. 
\end{enumerate}

\section{Example Domain-Expert Comments}
\label{sec:domain-expert-comments}
\begin{table*}[t!]
\centering
\scriptsize
\resizebox{\linewidth}{!}{%
\begin{tabular}{>{\centering\arraybackslash}p{0.02\linewidth} p{0.86\linewidth}}
\toprule
\textbf{\#} &   \textbf{Domain Expert Comment} \\
\midrule
\multicolumn{2}{l}{\textbf{Candidate discoveries identified by external reviewers, and confirmed by internal review}}    \\
\midrule
\rowcolor[HTML]{F3F3F3} 
1   &   \textbf{State Prediction Confidence:} Measuring prediction accuracy could be done using LLM-as-a-judge (e.g. have the model predict the observation, then have another LLM compare this generated observation to the gold observation, counting (perhaps by sentence, or by item) the number of things that are the same, and the number that are different, arriving at a score between 0-1 for each state prediction.  Similarly, do to the task well, the LLM doing the state prediction task should probably have at least the last 2-3 observations/actions in its prompt, to provide some context. \\%
2   &   \textbf{Accuracy vs Representational Expressivity:} No additional comments provided. \\%
\rowcolor[HTML]{F3F3F3} 
3   &   \textbf{Multi-Stage Environment Generation:} Solid idea -- try to build games incrementally rather than in one-shot, to see if that improves performance.  Doesn't mention where the source templates come from (presumably ideated from ByteSized32, so likely from that corpus/benchmark -- though it could also try to build them from scratch, or from a simple predefined template that it builds for this task, to make it easier).  It's also proposing to use a regex-based checker for game mechanics rather than the ByteSized32 evaluation methods -- that might work, or it might require an LLM-as-a-judge situation if the regex matching is not successful.  (Could include both in the evaluation, and compare them). \\%
4   &   \textbf{Combinatorial Optimization:} Could be interesting to see if an LLM can do this as well as a simple mathematical solver.  Should include a notion of tolerance (not in terms of the resistor tolerance, like 1\%, 5\%, etc., but in how close the value the different solvers create have to be to the real value -- otherwise some solutions may not be possible).  Should have a check that verifies the solutions (from the LLM, and other solvers) are within (say) 1\% or 5\% or 10\% of the expected value (or, could use all three of these, as a sort of graded accuracy metric). \\ %
\rowcolor[HTML]{F3F3F3} 
5   &   \textbf{Action Prediction:} Kind of makes sense, and would be interesting to see.  While the specification says to just provide a binary prediction (yes/no) as to whether the action will succeed (as well as the confidence score), it's not super clear what 'action will succeed' means.  Does it means the action will run in the interpreter? (in which case, it's not super interesting because, as long as the action is in the valid action list, it should run).  More interesting would be if it interpreted some signal that it worked (e.g. you can't cook a fridge or chop a pot, and the environment might say this, then (using a cheap LLM call), you might be able to interpret whether the observation returned after the action signified success or failure (e.g. 'you can't do that')).  But, extending this, it'd be interesting if it predicted more than binary success, but also did more of a state-prediction task -- e.g. predicting what the next observation will be, and then using an LLM to verify how much of it is essentially correct (perhaps proportion of sentences correct).  It'd need some number of steps of past history (say the last 1, 2, or 3 steps) to have a chance at doing this well. \\ %
6   &   \textbf{Graph Agent for Discovery:} Might be hard to get the DiscoveryWorld knowledge score working at the start (and extracting this coherently from the agent's memory) -- I'd focus on the DiscoveryWorld Task Completion and (more importantly) Task Process scores. \\ %
\rowcolor[HTML]{F3F3F3}
\midrule
\multicolumn{2}{l}{\textbf{Results rejected by either external reviewers, internal review, or both}}    \\
\midrule
\rowcolor[HTML]{F3F3F3}
7   &   \textbf{Social Graphs:} This could work -- but depends very much on the complexity and challenges required in interacting the social relationships.  It sounds like this proposes to create the benchmark rather than use an existing one -- so it would need to make sure that the interactions are interesting, reasonably complex, and non-trivial to navigate.  It'd also need some clear measure of evaluating an agent's performance -- it's not clear what 'accuracy of relationship-based decisions' is or how it would be measured. \\ %
15   &   \textbf{Planning Agent:} Mostly makes sense, but one of its assumptions (focusing on get/put/cook recipes) isn't possible, it'd have to change this -- there's no way of limiting what actions need to be used.  Also it should use the task score, not task completion rate. Most agents do not complete any tasks, but the task score is a partial score between 0 and 1 that is often non-zero if an agent makes task progress. \\ %
\rowcolor[HTML]{F3F3F3}
9   &   \textbf{Spatial Agent:} Makes a lot of sense, and you'd expect this to work.  Somewhat related to other agents (though I'm not sure any have tried augmenting ReAct in this way, or on this environment).  Should use the partial task score instead of task completion rate as a measure of success -- the tasks are hard and most agents don't complete them, but the partial task score gives a score between 0-1 that measures partial progress. \\    %
10   &   \textbf{Action History:} Makes sense.  There have been a lot of similar ideas generated, the one that makes this one more viable is that it's not just tracking successful actions in isolation, but considering the *context* in which they occurred.  Text games generally require long action sequences, where each action is taken at the appropriate time, when all the right conditions are met.  Taking the context into account should help it figure out when it's appropriate to take a particular action.  Progress should be measured using the Task Score (a measure of partial progress), not the Task Success Rate, since task success is rare with most agents in these environments. \\ %
\rowcolor[HTML]{F3F3F3}
11   &   \textbf{Template Agent:} Makes sense, but (1) should use increasing partial task score (0-1), rather than task success/completion, as a signal -- since this environment is hard, and task success is rare. \\    %
12   &   \textbf{Graph Verification Agent:} No additional comments provided. \\ %
\rowcolor[HTML]{F3F3F3}
13   &   \textbf{Affordance Agent:} Makes sense -- uses an LLM to predict affordances, then act based on those affordances.  Perhaps could augment a ReAct agent with 3 steps (affordances, think, act) rather than just the normal 2 (think/act).  Should measure performance in terms of the task score, rather than task success (since task success is rare).  Could use the 'find living thing' subtask (one of the easiest ones) as an additional task to try. \\ %
14   &   \textbf{WordNet:} Super simple baseline agent. One might expect it to outperform a random baseline.  It might even outperform a ReAct agent on this task, especially if paired with aspects that allow the agent to explore the environment.  Should likely be run on very simple (e.g. 3 room maximum) environments. \\ %
\rowcolor[HTML]{F3F3F3}
15   &   \textbf{Metaphor Graphs:} At first glance it's hard to see how metaphors would be useful here, but the suggested operationalization (e.g. 'what functional similarities exist between X and Y in a cooking context?') might help it better organize the graph into categories of objects.  The ``project supervisor'' ratings (i.e. manual human ratings) should likely not be included, since this requires human ratings, and interrupts the automatic flow of running the experiment. \\ %
16   &   \textbf{Goal Tracking:} It might work, though it doesn't mention a base agent (like ReAct) to augment with the goals.  It's good that it mentions using task score (rather than task completion) as a metric, since task completion is often zero for these hard tasks, where as task score is often non-zero if the agent is making some progress. \\ %
\rowcolor[HTML]{F3F3F3}
17   &   \textbf{Container Agent:} It's specifically focused on building a graph of relevant container relationships, which at first seemed uninteresting, but now that I think about it, basic e.g. ReAct agents tend to struggle with finding ingredients -- so having a graph of where they tend to be could help it.  Same for tools it needs (e.g. cooking implements, recipe book, knife for chopping, etc.).  Presumably the graph would be included in the ReAct agent prompt. The metric should not be task completion (since task success is hard and rarely non-zero on this task), but rather the task score, which provides a partial measure of task progress (with a value between zero and one). \\ %
18   &   \textbf{Template-based Environment Creation:} Mostly makes sense, but one of its assumptions (focusing on get/put/cook recipes) isn't possible, it'd have to change this -- there's no way of limiting what actions need to be used.  Also it should use the task score, not task completion rate. Most agents do not complete any tasks, but the task score is a partial score between 0 and 1 that is often non-zero if an agent makes task progress. \\ %
\rowcolor[HTML]{F3F3F3}
19   &   \textbf{Graph Metric:} Neat idea (and, very different than many of the others).  Would benefit from using vector similarity. \\%

\bottomrule
\end{tabular}%
}
\caption{\footnotesize Domain expert comments provided at the ideation stage, for each of the 19 potential discoveries from Table~\ref{tab:discoveries}. \label{tab:domain-expert-comments} }
\end{table*}

The domain expert comments are appended to each selected idea before the planning stage, typically to correct minor issues (such as selecting a better metric), or clarify portions of the idea.  Pilot experiments without these comments showed that the experiments generally still work, but have less-strong conclusions (e.g. a less robust metric will be used, such a direct string matching instead of a more robust LLM-as-a-judge metric, limiting what can be claimed), or may not find results at all (e.g. the system may wish to measure an agent's success only by task completion, which is often zero for most agents and environments -- whereas instead using the normalized partial progress scores offered by most environments is standard practice).  To highlight the limited nature of these comment, the expert comments for each of the 19 discoveries in Table~\ref{tab:discoveries} are provide in Table~\ref{tab:domain-expert-comments}

\section{Example Idea, Comments, and Plan}
\label{sec:idea-operationalization-example}

An example automatically generated idea, set of domain-expert comments, and the resulting generated plan from the combination of the two are provided in Table~\ref{tab:ideation-operationalization-example}.

\begin{table*}[t!]
\centering
\scriptsize
\begin{tabular}{p{0.15\linewidth} p{0.80\linewidth}}
\toprule
\rowcolor{gray!20} \textbf{Field} & \textbf{Content} \\
\midrule
\multicolumn{2}{l}{\textit{Ideation}}\\
\midrule
name & simulation-confidence-analysis \\[0.5em]
long\_description & Study whether LLMs can accurately assess their confidence in state predictions, and whether this confidence correlates with actual accuracy. This could enable more reliable simulation by identifying when predictions are likely to be incorrect. \\[1em]
short\_description & Investigate LLM ability to assess confidence in state predictions and correlation with accuracy. \\[1em]
hypothesis & LLM confidence scores will correlate with prediction accuracy, allowing for identification of potentially incorrect predictions. \\[1em]
variables & Independent variables: State complexity, Game type, Property type. Dependent variables: Prediction accuracy, Confidence score. Control: Same LLM, same states, same examples. \\[1em]
metric & Correlation between confidence scores and accuracy. Precision/recall for identifying incorrect predictions using confidence thresholds. \\[1em]
pilot & Test on simple CookingWorld scenarios, focusing on boolean property predictions with confidence scores. \\[1em]
example\_prompt & Create an experiment to analyze LLM confidence in state predictions. Use TextWorldExpress to generate 200 state transitions. For each prediction, prompt GPT-4 to provide both the predicted state and a confidence score (0--100) for each property change. Log all predictions, confidence scores, and ground truth. Calculate correlation between confidence and accuracy. Generate ROC curves for using confidence to predict correctness. Use bootstrap resampling to compute confidence intervals. Create visualizations showing the relationship between confidence and accuracy across different property types. \\[1em]
\midrule
\multicolumn{2}{l}{\textit{Human Rating}}\\
\midrule
human\_rating & potentially feasible \\[1em]
human\_notes & Measuring prediction accuracy could be done using LLM-as-a-judge (e.g. have the model predict the observation, then have another LLM compare this generated observation to the gold observation, counting (perhaps by sentence, or by item) the number of things that are the same, and the number that are different, arriving at a score between 0--1 for each state prediction). Similarly, the LLM doing the state prediction task should probably have at least the last 2--3 observations/actions in its prompt to provide context. \\[1em]
conditioning\_text & Please use \texttt{gpt-4o-mini} for all LLM calls, because it's fast and inexpensive. \\[1em]
\midrule
\multicolumn{2}{l}{\textit{Operationalization}}\\
\midrule
operationalization & 
Please create an experiment to analyze LLM confidence in state predictions in TextWorldExpress, implementing the following specifications: 
\textbf{EXPERIMENT MODES AND SCOPE:} Implement a global variable \texttt{PILOT\_MODE} that can be set to one of: 'MINI\_PILOT', 'PILOT', or 'FULL\_EXPERIMENT'. For each mode: \newline
\quad \texttt{MINI\_PILOT}: 3 episodes of CookingWorld, 10 steps each (training set); \newline
\quad \texttt{PILOT}: 20 episodes of CookingWorld, 25 steps each (training set); \newline
\quad \texttt{FULL\_EXPERIMENT}: 200 episodes, 50 steps each (balanced across train/dev/test sets). \newline
\textbf{ENVIRONMENT SETUP:} Use the TextWorldExpress API to generate CookingWorld scenarios with simple environments (3 rooms, no doors, 2 ingredients, 2 distractor items). Record the current state, action, and next state at each step. \newline
\textbf{LLM CONFIGURATION:} Use \texttt{gpt-4o-mini} for all LLM calls. The state prediction prompt should include the last 2 observations, the current action, a request for the next state prediction, and a confidence score (0--100) for each property change. \newline
\textbf{DATA COLLECTION PROCEDURE:} For each step, retrieve the current state and action; prompt the LLM for a state prediction with confidence scores; retrieve the actual next state; and use an LLM-as-a-judge to score prediction accuracy (score: 0--1 per property change). \newline
\textbf{DATA ANALYSIS:} For each episode, calculate the correlation between confidence scores and accuracy, and generate scatter plots, ROC curves, and visualizations of the confidence--accuracy relationship. \newline
\textbf{LOGGING AND OUTPUT:} Log all raw data (states, predictions, confidence scores, accuracy scores), generate summary statistics, and create plots (scatter plot, ROC curves, box plots). \newline
\textbf{EXECUTION FLOW:} Run \texttt{MINI\_PILOT} first; if successful, run \texttt{PILOT} and then stop. \newline
\textbf{SUCCESS CRITERIA:} Achieve clean execution in \texttt{MINI\_PILOT} and find meaningful correlation patterns in \texttt{PILOT}, with statistical significance verified via bootstrap resampling. \\[1em]

code\_examples & TextWorldExpress API Example, Non-parametric Bootstrap Resampling, Logger/Debugging, MatPlotLib Line Plot, LLM example through proxy server \\
\bottomrule
\end{tabular}
\caption{\footnotesize An example idea, the domain expert comments, and the generated plan (produced by using both the idea and comments). }
\label{tab:ideation-operationalization-example}
\end{table*}

\setminted{fontsize=\scriptsize, breaklines=true, breaksymbolleft={\textcolor{lightgray}{\textbackslash}}}
\clearpage
\section{Prompts}
\label{sec:appendix-prompts}
Core prompts are provided below.

\clearpage
\label{sec:prompt-ideation}
\onecolumn
\begin{tcolorbox}[colback=ideationred!10, colframe=ideationred!20, enhanced, breakable, listing only, width=\textwidth, coltitle=black, title=Ideation Prompt]
\begin{minted}{text}
You are ScientistGPT, the most advanced automated scientific model in the world. You can use your enormous intellect to solve any problem, and the solutions to these problems may help improve our knowledge of how the world works, which is a noble and important goal.
You are currently working on the following task: Generating new research ideas/ideas for new experiments to run.
The goal of running the experiments is to generate novel, interesting, and (ideally) high-impact scientific results.
Below is a set of scientific research papers (expressed as their Latex source).
Your task is to come up with new research ideas, and follow-on research ideas, based on the research questions, research programs, hypotheses, operationalizations of experiments, or any other information provided in these papers.
You are asked to come up with 5 ideas.
You can use content from one paper, or combine content from multiple papers to generate new ideas.
The ideas you generate can be highly novel inspired by the research in the papers below, or they can be incremental follow-on ideas based on the papers below. The most important thing is that we're doing good, reasoned, and potentially impactful/useful science.
Some recipes for ideation you might use are the following:
1. **Filling the gaps**: Identify gaps (high-level or low-level) in the research programs below, and come up with ideas to fill those gaps.
2. **Abstractive**: Abstract the research programs below to a higher level, and come up with new ideas based on those abstractions.
3. **Combining ideas**: Combine ideas from different research programs below to come up with new ideas.
4. **Extending ideas**: Extend the ideas from the research programs below to come up with new ideas.
5. **Challenging assumptions**: Challenge the assumptions made in the research programs below, and come up with new ideas based on those challenges.
6. **What happens if**: Come up with ideas that ask what happens if you change key/important parts of the research programs below.
7. etc.

After reading the research papers (and their implicit or explicit Research Programs, Hypotheses, and Operationalizations of Experiments contained within the papers) below, you will be asked to come up with a list of new research ideas (which can be highly novel or incremental follow-on ideas).
As a strategy, you can try coming up with one idea for *each* of the methods above (i.e. filling the gaps, abstractive, combining ideas, extending ideas, challenging assumptions, etc.), or subsampling this if you need to generate fewer ideas.
The response format (JSON) is below:
```json
[   # List of research ideas
    {    # Research Idea 1
         "research_idea_name": "A 2-3 word name of this research idea, hypthen-separated (e.g. my-research-idea)",
         "research_idea_long_description": "A long (e.g. ~50-100 word) description of the research idea, and what it's investigating.",
         "research_idea_short_description": "A short (e.g. ~20 word) high-level description of the research idea, and what it's investigating.",
         "research_idea_hypothesis": "What is the hypothesis of the research idea?  What are you trying to prove or disprove?",
         "research_idea_variables": "What are the main variables involved in investigating this research program?  What variables are held constant, and what variables are manipulated?",
         "research_idea_metric": "What is the main metric that will be used to evaluate the success of this research idea?  How will we know if the idea works or not?  How will partial performance be measured?",
         "research_baselines": "If your system is an experimental system, what baselines will you compare against?  I.e. if you're creating a (a) new method based on an old method, or a (b) modification of an existing method, then you should probably compare to the old method or the existing method.  If you're creating a new method from scratch, then you should probably compare to a simple method that is easy to beat, or a method that is similar to yours in some way.",
         "research_idea_pilot": "What's the simplest version of this that can be tested, before running a more expensive version?  Usually this is a full (or reasonably full) method, but on a small subset of the input data.",
         "research_idea_design_prompt": "Provide a detailed design of the experiment here, with enough detail that it can be implemented by a student-level practitioner (which in actuality, is an automated experiment building system).  This is the only text they will be provided to build the experiment, so be specific.  DO NOT JUST GIVE HIGH-LEVEL DESCRIPTIONS (like 'implement the experiment') because this is useless.  This should minimally include at least: (1) *Detailed* mid-level descriptions of what is to be implemented, including any algorithms (designed at a high or low level), (2) Detailed descriptions of what data to use, in the context of a pilot experiment, (3) Detailed descriptions of what output to generate, how to save it (in the context of maximum utility for follow-on experiments), and how to evaluate and report the results."
         "research_idea_codeblocks": ["A list of existing codeblocks from the codeblock library that this idea is likely to use"],
         "research_idea_required_code_and_resources": [ # An EXHAUSTIVE list of ALL required CODE, RESOURCES, MODELS, etc. mentioned in this ENTIRE RESERACH IDEA.  CRITICALLY IMPORTANT, USED TO DETERMINE FEASIBILITY! If it's mentioned above, it ABSOLUTELY NEEDS to be here!
             {"name": "example short name", "description": "a short example description of the code or resource that is needed", "where": "one of: `existing codeblock`, `external`, or `build`", "effort": "one of: `minor`, `moderate`, or `major`"}, # `where` refers to where the code/resource/model comes from (an existing codeblock template, an external source that can be retrieved, or whether we need to build it for this work.  `effort` refers to how much effort that process will take: `minor` (e.g. small modifications/trivial code), `moderate` (a good amount of work), and `major` for large and/or high-difficulty volumes of code.
             {"name": "ReAct baseline", "description": "A ReAct baseline (targeted for use on Benchmark XYZ)", "where": "existing codeblock", "effort": "minor"}, # `existing codeblock` because there's an existing codeblock covering a ReAct baseline, and `minor` because this is just using the existing codeblock.
             {"name": "Modified ReAct baseline", "description": "The proposed modified ReAct model", "where": "existing codeblock", "effort": "moderate"}, # `existing codeblock` because there's an existing codeblock covering a ReAct baseline, and `moderate` because this is proposing non-trivial (but not huge) modifications to that agent, so it will take some work to build.
             {"name": "Benchmark X", "description": "The primary benchmark used to evaluate the models", "where": "existing codeblock", "effort": "minor"},  # `existing codeblock` because this benchmark has an existing codeblock covering it, that just needs to be directly applied 
             {"name": "Benchmark Y", "description": "The secondary (transfer) benchmark used to evaluate the models", "where": "external", "effort": "moderate"},    # `external` because there's no existing codeblock for this specific benchmark. `moderate` because this isn't a popular benchmark available on e.g. huggingface, so it will likely take a bit of work to find/download/load it.
             {"name": "LLM interface", "description": "The interface to prompt the LLM for the agents", "where": "existing codeblock", "effort": "minor"}, # The agents need LLM calls. This is an existing codeblock with minor modifications (if any).
             {"name": "gpt-4o model", "description": "The gpt-4o model available from the OpenAI API", "where": "existing codeblock", "effort": "minor"}, # The base model to use for the agents. The LLM codeblock covers using it, so we don't need to download it, and it should be low effort.
             {"name": "Prior Agent ABC", "description": "An existing agent described in the paper, to serve as a secondary baseline", "where": "external", "effort": "major"}, # `external` because we don't have a codeblock for it and it's something someone else published on github. `moderate` since it's usually a fair amount of work getting someone else's agent working.
             {"name": "Fancy New Agent++", "description": "A new agent proposed in this work integrating ...", "where": "build", "effort": "major"}, # `build` because we have to largely build it from scratch, and `major` because it's a fairly complex new agent algorithm that is likely to take a lot of work to build.
             {"name": "Bootstrap resampling", "description": "The bootstrap resampling technique for comparing the performance of two models", "where": "existing codeblock", "effort": "minor"}, # Already compltely covered in an existing codeblock, we just have to use it
             {"name": "New dataset collection", "description": "A new dataset collection procedure for collecting data for the agents through web scraping", "where": "build", "effort": "major"}, # `build` because we have to build it from scratch, and `major` because it's a fairly complex new data collection procedure that is likely to take a lot of work to build and debug.
             # ... More code/resources/models/etc, if any
         "research_idea_external_requirements": ["An exhaustive list of libraries or packages that may be required. Format: `python/apt package name (very short description of need)`", "transformers (for XYZ)", "scikit-learn (for ABC)", ...]
     }, 
     ... # More research ideas
]
```

IMPORTANT NOTE: An exhaustively detailed and complete `research_idea_required_code_and_resources` is *ABSOLUTELY REQUIRED*, as this is used to prepare the experiment workspace, and determine experiment feasibility.  A poorly or incorrectly documented `research_idea_required_code_and_resources` for an idea is a major failure, as it will waste a large amount of resources (time/money/etc) on ideas that may be unlikely to have the resources they need to succeed.

NOTE: Below is a simple example `research_idea_design_prompt` for a hypothetical example research idea:
```
Please create an agent that automatically builds an informative, useful knowledge graph from exploring its environment. The knowledge graph should be expressed as triples, i.e. subject-relation-object, and stored in DOT/Graphviz format. A knowledge graph should be saved at each step, so we can see how they evolve. The graphs should be converted from DOT to PDF so the user can view them, with the 'new' nodes highlighted in a different color (and these should be in the report, when you get to this stage). Please test this on CookingWorld, using the default CookingWorld environment parameters (except 3 rooms, and no doors). The base model should be `gpt-4o-mini`. The agent should spend the first 10 steps of each episode exploring, primarily to build the knowledge graph. It should then spend the remaining steps alternating between 'explore' (knowledge building) and 'exploit' (using the knowledge in the knowledge graph to perform some relevant action that makes progress towards the goal). The agent should use the first 2 parametric variations (i.e. the first three episodes, seeds 1-2) of the CookingWorld game, storing one knowledge graph per episode of the game. The maximum steps per episode should be 40. The full trajectory (i.e. observation, score, possible valid actions, chosen action at each step) should be in the log file.
```

You are asked to generate new research ideas that are *conditioned*/*related to* the kinds of codeblocks that the automated experiment builder has available in the codeblock library. Here are high-level summaries of the code templates available in the experiment builder:
```
<Insert codeblock summaries here>
```

Existing Research Papers, from which you should consider their (implicitly or explicitly stated) Research Programs, Hypotheses, and Operationalizations of Experiments:
Example Paper 1:
```
<Insert Latex of Paper 1 Here>
```

Example Paper 2:
```
<Insert Latex of Paper 2 Here>
```


Please generate a list of new research ideas (which can be highly novel or incremental follow-on ideas). The most important thing is that we're doing good, reasoned, and potentially impactful/useful science.
After reading the research papers (and their implicit or explicit Research Programs, Hypotheses, and Operationalizations of Experiments contained within the papers) below, you will be asked to come up with a list of new research ideas (which can be highly novel or incremental follow-on ideas).
You are asked to come up with 5 ideas.
As a strategy, you can try coming up with one idea for *each* of the methods above (i.e. filling the gaps, abstractive, combining ideas, extending ideas, challenging assumptions, etc.), or subsampling this if you need to generate fewer ideas.
The response format (JSON) is below:
```json
[   # List of research ideas
    {    # Research Idea 1
         "research_idea_name": "A 2-3 word name of this research idea, hypthen-separated (e.g. my-research-idea)",
         "research_idea_long_description": "A long (e.g. ~50-100 word) description of the research idea, and what it's investigating.",
         "research_idea_short_description": "A short (e.g. ~20 word) high-level description of the research idea, and what it's investigating.",
         "research_idea_hypothesis": "What is the hypothesis of the research idea?  What are you trying to prove or disprove?",
         "research_idea_variables": "What are the main variables involved in investigating this research program?  What variables are held constant, and what variables are manipulated?",
         "research_idea_metric": "What is the main metric that will be used to evaluate the success of this research idea?  How will we know if the idea works or not?  How will partial performance be measured?",
         "research_baselines": "If your system is an experimental system, what baselines will you compare against?  I.e. if you're creating a (a) new method based on an old method, or a (b) modification of an existing method, then you should probably compare to the old method or the existing method.  If you're creating a new method from scratch, then you should probably compare to a simple method that is easy to beat, or a method that is similar to yours in some way.",
         "research_idea_pilot": "What's the simplest version of this that can be tested, before running a more expensive version?  Usually this is a full (or reasonably full) method, but on a small subset of the input data.",
         "research_idea_design_prompt": "Provide a detailed design of the experiment here, with enough detail that it can be implemented by a student-level practitioner.  This is the only text they will be provided to build the experiment, so be specific.  This should minimally include at least: (1) Detailed descriptions of what is to be implemented, including any algorithms (designed at a high or low level), (2) Detailed descriptions of what data to use, in the context of a pilot experiment, (3) Detailed descriptions of what output to generate, how to save it (in the context of maximum utility for follow-on experiments), and how to evaluate and report the results."
         "research_idea_codeblocks": ["A list of existing codeblocks from the codeblock library that this idea is likely to use"],
         "research_idea_required_code_and_resources": [ # An EXHAUSTIVE list of ALL required CODE, RESOURCES, MODELS, etc. mentioned in this ENTIRE RESERACH IDEA.  CRITICALLY IMPORTANT, USED TO DETERMINE FEASIBILITY! If it's mentioned above, it ABSOLUTELY NEEDS to be here!
             {"name": "example short name", "description": "a short example description of the code or resource that is needed", "where": "one of: `existing codeblock`, `external`, or `build`", "effort": "one of: `minor`, `moderate`, or `major`"}, # `where` refers to where the code/resource/model comes from (an existing codeblock template, an external source that can be retrieved, or whether we need to build it for this work.  `effort` refers to how much effort that process will take: `minor` (e.g. small modifications/trivial code), `moderate` (a good amount of work), and `major` for large and/or high-difficulty volumes of code.
             {"name": "ReAct baseline", "description": "A ReAct baseline (targeted for use on Benchmark XYZ)", "where": "existing codeblock", "effort": "minor"}, # `existing codeblock` because there's an existing codeblock covering a ReAct baseline, and `minor` because this is just using the existing codeblock.
             {"name": "Modified ReAct baseline", "description": "The proposed modified ReAct model", "where": "existing codeblock", "effort": "moderate"}, # `existing codeblock` because there's an existing codeblock covering a ReAct baseline, and `moderate` because this is proposing non-trivial (but not huge) modifications to that agent, so it will take some work to build.
             {"name": "Benchmark X", "description": "The primary benchmark used to evaluate the models", "where": "existing codeblock", "effort": "minor"},  # `existing codeblock` because this benchmark has an existing codeblock covering it, that just needs to be directly applied 
             {"name": "Benchmark Y", "description": "The secondary (transfer) benchmark used to evaluate the models", "where": "external", "effort": "moderate"},    # `external` because there's no existing codeblock for this specific benchmark. `moderate` because this isn't a popular benchmark available on e.g. huggingface, so it will likely take a bit of work to find/download/load it.
             {"name": "LLM interface", "description": "The interface to prompt the LLM for the agents", "where": "existing codeblock", "effort": "minor"}, # The agents need LLM calls. This is an existing codeblock with minor modifications (if any).
             {"name": "gpt-4o model", "description": "The gpt-4o model available from the OpenAI API", "where": "existing codeblock", "effort": "minor"}, # The base model to use for the agents. The LLM codeblock covers using it, so we don't need to download it, and it should be low effort.
             {"name": "Prior Agent ABC", "description": "An existing agent described in the paper, to serve as a secondary baseline", "where": "external", "effort": "major"}, # `external` because we don't have a codeblock for it and it's something someone else published on github. `moderate` since it's usually a fair amount of work getting someone else's agent working.
             {"name": "Fancy New Agent++", "description": "A new agent proposed in this work integrating ...", "where": "build", "effort": "major"}, # `build` because we have to largely build it from scratch, and `major` because it's a fairly complex new agent algorithm that is likely to take a lot of work to build.
             {"name": "Bootstrap resampling", "description": "The bootstrap resampling technique for comparing the performance of two models", "where": "existing codeblock", "effort": "minor"}, # Already compltely covered in an existing codeblock, we just have to use it
             {"name": "New dataset collection", "description": "A new dataset collection procedure for collecting data for the agents through web scraping", "where": "build", "effort": "major"}, # `build` because we have to build it from scratch, and `major` because it's a fairly complex new data collection procedure that is likely to take a lot of work to build and debug.
             {"name": "Cohen's Kappa", "description": "The Kappa measure of interannotator agreement for the dataset (use the `sklearn` library implementation)", "where": "build", "effort": "minor"}, # Fairly straightforward use of this library, so it's a minor effort.
             {"name": "Rouge score", "description": "The Rouge score for evaluating the quality of the generated text (use the `rouge-score` library implementation)", "where": "existing codeblock", "effort": "minor"}, # Already covered in an existing codeblock, so it's a minor effort.
             # ... More code/resources/models/etc, if any
         "research_idea_external_requirements": ["An exhaustive list of libraries or packages that may be required. Format: `python/apt package name (very short description of need)`", "sklearn (for kappa)", "rouge-score (for rouge score)", ...]
     }, 
     ... # More research ideas
]
```

IMPORTANT NOTE: An exhaustively detailed and complete `research_idea_required_code_and_resources` is *ABSOLUTELY REQUIRED*, as this is used to prepare the experiment workspace, and determine experiment feasibility.  A poorly or incorrectly documented `research_idea_required_code_and_resources` for an idea is a major failure, as it will waste a large amount of resources (time/money/etc) on ideas that may be unlikely to have the resources they need to succeed.

NOTE: Below is a simple example `research_idea_design_prompt` for a hypothetical example research idea:
```
Please create an agent that automatically builds an informative, useful knowledge graph from exploring its environment. The knowledge graph should be expressed as triples, i.e. subject-relation-object, and stored in DOT/Graphviz format. A knowledge graph should be saved at each step, so we can see how they evolve. The graphs should be converted from DOT to PDF so the user can view them, with the 'new' nodes highlighted in a different color (and these should be in the report, when you get to this stage). Please test this on CookingWorld, using the default CookingWorld environment parameters (except 3 rooms, and no doors). The base model should be `gpt-4o-mini`. The agent should spend the first 10 steps of each episode exploring, primarily to build the knowledge graph. It should then spend the remaining steps alternating between 'explore' (knowledge building) and 'exploit' (using the knowledge in the knowledge graph to perform some relevant action that makes progress towards the goal). The agent should use the first 2 parametric variations (i.e. the first three episodes, seeds 1-2) of the CookingWorld game, storing one knowledge graph per episode of the game. The maximum steps per episode should be 40. The full trajectory (i.e. observation, score, possible valid actions, chosen action at each step) should be in the log file.
```
NOTE: You should try to define important terms, as the paper is likely to be unavailable to the automated experiment builder, only the information you provide will be.  Similarly, acronyms can be used, but they should be defined on their first use.

Your JSON response must be between code blocks (```).  You can write any other text you wish before or after (such as if you want to describe the research programs, hypotheses, and/or operationalizations of experiments in the papers), but only JSON text between a single set of codeblocks (```) will be able to be automatically extracted and used.
Remember that your research ideas should be ACTUALLY IMPLEMENTABLE by being conditioned on the kinds of codeblocks that the automated experiment builder has available.  Your templates and operationalizations should particularly emphasize existing codeblocks in the experiment builder.
Similarly, while you can use external libraries or packages, you may wish to minimize the use of these, as they may not be available to the automated experiment builder, or (more likely) it may not be fluent in their use.
\end{minted}
\end{tcolorbox}
\captionof{listing}{Ideation Prompt}

\clearpage
\label{sec:appendix-operationalization-prompt}
\onecolumn
\begin{tcolorbox}[colback=planningyellow!10, colframe=planningyellow!20, enhanced, breakable, listing only, width=\textwidth, coltitle=black, title=Planning Prompt]
\begin{minted}{text}
You are ScientistGPT, the most advanced automated scientific model in the world. You can use your enormous intellect to solve any problem, and the solutions to these problems may help improve our knowledge of how the world works, which is a noble and important goal.
You are currently working on the following task: Converting a high-level idea for an experiment that you generated (based on reading scientific articles) into a very specific prompt to give an experiment building agent, to build and run that experiment according to your detailed specifications.
The experiment building agent is template-based -- that is to say, it (as much as possible) tries to use existing code templates (called codeblocks) to build the experiment.  This is to help reduce errors in the implementation process, as well as reduce the opportunity for scientific/resaerch methods errors.
Below is the high-level experiment idea that you generated.  Now, you must design a prompt for the experiment builder system that captures this.

Your experiment idea to convert into a prompt for the experiment builder is the following:
(Note that information provided in the idea here may not be completely accurate or usable as-is -- for example, operationalizing the idea may require more or different codeblock templates than what are mentioned in the idea below (if the idea even suggests code blocks).  Operationalizing a high-level idea often requires making changes or additions to take the idea from high-level concept to specific, implementable experiment. Please use your best judgement.)
```
<Insert research idea here>
```

In addition, you are asked to use the following SPECIAL CONDITIONING INSTRUCTIONS, that usually help re-scope the experiment to be more implementable, scope the experiment to be more towards user goals, and/or help reduce errors in the implementation process:
```
<Insert any special conditioning instructions, such as:>
Please use `gpt-4o-mini` for all LLM calls, because it's fast and inexpensive.
```

Your output must contain two keys: `prompt` and `codeblocks`.  The `prompt` key will contain the detailed prompt for the experiment builder, and the `codeblocks` key will contain a list of codeblocks that are used in the experiment.
Consequences of errors in `prompt` and `codeblocks`:
1. If the prompt is not detailed or useful, the experiment builder may not be able to build the experiment (but it will still try), and this will waste a lot of time and resources.
2. If the required codeblocks are not included, the experiment builder is highly unlikely to build the experiment successfully (but it will still try), and this will waste a lot of time and resources.

Here are two (very simple, very rough) examples of what a prompt might look like for a very simple, very hypothetical, toy experiment:
Example Prompt Generation #1:
```
{
    "prompt": "Please investigate the effect of implementing a ReAct agent with and without a small difference. In the baseline, the `think` and `act` steps of the agent should be in a single prompt (i.e. a single LLM call). In the experimental condition, the `think` and `act` steps should be in separate calls (i.e. it thinks, then it acts based on the thought). Please test this on CookingWorld, using the default CookingWorld environment parameters (except 3 rooms, and no doors). The base model should be `gpt-4o-mini`. The agent should use the first 5 parametric variations (i.e. the first five episodes, seeds 1-5) of the CookingWorld game, and end after this, report the score/success of each episode, and final average score. The maximum steps per episode should be 25. The full trajectory (i.e. observation, score, possible valid actions, chosen action at each step) should be in the log file. The results file should include number of steps per episode, as well as an average of this. Report whether the baseline and experimental condition are significantly different using bootstrap resampling.",
    "codeblocks": ["Logger/Debugging", "LLM example through proxy server", "ReAct Agent Example", "TextWorldExpress API Example", "Non-parametric Bootstrap Resampling"]
}
```

Example Prompt Generation #2:
```
{
    "prompt": "Please create an agent that automatically builds an informative, useful knowledge graph from exploring its environment. The knowledge graph should be expressed as triples, i.e. subject-relation-object, and stored in DOT/Graphviz format. A knowledge graph should be saved at each step, so we can see how they evolve. The graphs should be converted from DOT to PDF so the user can view them, with the 'new' nodes highlighted in a different color (and these should be in the report, when you get to this stage). Please test this on CookingWorld, using the default CookingWorld environment parameters (except 3 rooms, and no doors). The base model should be `gpt-4o-mini`.  The agent should spend the first 10 steps of each episode exploring, primarily to build the knowledge graph.  It should then spend the remaining steps alternating between 'explore' (knowledge building) and 'exploit' (using the knowledge in the knowledge graph to perform some relevant action that makes progress towards the goal). The agent should use the first 2 parametric variations (i.e. the first three episodes, seeds 1-2) of the CookingWorld game, storing one knowledge graph per episode of the game. The maximum steps per episode should be 40. The full trajectory (i.e. observation, score, possible valid actions, chosen action at each step) should be in the log file.
    "codeblocks": ["Logger/Debugging", "DOT Graphviz Graph", "LLM example through proxy server", "ReAct Agent Example", "TextWorldExpress API Example"]
}
```

*Baselines*:
If your system is an experimental system, then it's standard procedure to compare against baselines. Baselines are usually one of the following:
1. If you're creating a new method based on an old method, or a modification of an existing method, then you should probably compare to the old method or the existing method.
2. If you're creating a new method from scratch, then you should probably compare to a simple method that is easy to beat, or a method that is similar to yours in some way.
3. Sometimes, you might compare to both of the above. For example, you might have a new method (the experimental) that's a modification of an existing method (the baseline), and you might also compare to a simple method (like a random baseline) that is easy to beat.
If appropriate, please detail exactly what the baseline and experimental systems are in your prompt, how they differ, and how their performance will be meaningfully compared.

You are asked to generate an experiment prompt that is conditioned on the actual code templates available in the system, as much as possible, to help reduce the errors.  Here is a high-level summary of the codeblocks:
```
<Insert high-level summaries of codeblocks in codeblock library>
```

The following codeblocks are mentioned in the research idea, that may help you generate your experiment design prompt:

<Insert listings of codeblocks mentioned in the idea>

```


Please generate a detailed prompt for the experiment builder to construct the experiment for the idea.  That idea again is:
```
<Insert research idea here>
```

Your output must be a JSON dictionary containing two keys: `prompt` and `codeblocks`.  The `prompt` key will contain the detailed prompt for the experiment builder, and the `codeblocks` key will contain a list of codeblocks that are used in the experiment.
Your output must be a JSON dictionary between code blocks (```).  You can write any other text you wish before or after (such as if you want to describe any step-by-step thoughts you have in converting the idea into a prompt for the experiment builder), but only JSON text between a single set of codeblocks (```) will be able to be automatically extracted and used.
For example:
```
{
     "prompt": "The detailed prompt for the experiment builder goes here.",
     "codeblocks": ["List of codeblocks used in the experiment.  They must exactly match the codeblock names. If zero codeblocks are required, you must output a blank list here."]
}
```

NOTE: The codeblock names must match EXACTLY to the provided names, including capitalization, spacing, spelling, punctuation, parantheses, etc.  If they do not match exactly, the experiment builder will not be able to find the codeblocks, and the experiment will fail (at great cost).

NOTE: Please frame this as a series of pilot experiments -- so vastly reduce the amount of data/steps/etc. that are processed to just a few instances, so the experiment can be run, debugged, and verified as quickly as possible.  More details on the pilot experiment setting:
 - There should be a global variable in your code (PILOT_MODE:str) with three possible settings: `MINI_PILOT`, `PILOT`, or `FULL EXPERIMENT`.
 - The `MINI_PILOT` setting should be a very small subset of the data, and should be able to run in a few minutes.  The purpose is for fast debugging and verification of the code. For example, for question answering tasks, this might be 10 questions.  For agent tasks, this might be 2-3 episodes at 10-20 steps each.  The questions/episodes should come from the training set.
 - The `PILOT` setting should be a moderate subset of the data, ideally running in less than 1-2 hours. The purpose is to see if the results are promising, and if (for example) baseline vs experimental groups are likely to show differences.  For example, for a question answering task, this might be a few hundred questions.  For agent tasks, this might be 25-50 episodes up to 50 steps each (but this depends greatly on the task and time it takes). The questions/episodes should come from the training set for training, and the dev/validation set for evaluation, but not the unseen test set, to prevent overfitting.
 - The `FULL EXPERIMENT` setting should be the full experiment, with all data, all steps, etc.  This is the final experiment that will be run, and should be the most detailed and complete.  Training data should come from the training set.  Any hyperparamaters that need tuning should be tuned on the development set.  The experiment should be evaluated on the test set.
 - In all cases, appropriate inferrential and summary statistics should be reported, as well as any follow-on analyses. The difference between pilot levels is simply of scale, not of quality.
 - Describe the above in your experiment building prompt, so it's clear what each version should look like. - In the experiment prompt, say that it should run the MINI_PILOT first, then if everything looks good, the PILOT.  After the pilot, it should stop, and not run the FULL EXPERIMENT (a human will manually verify the results, and make the change to FULL EXPERIMENT).Please generate your JSON output now. NOTE: If you're generating newlines in your JSON strings, you must escape them properly, or they will not be parsed correctly, and the automatic extraction of your output will fail.
\end{minted}
\end{tcolorbox}
\captionof{listing}{Planning Prompt}

\clearpage
\label{sec:prompt-experiment-builder}
\onecolumn
\begin{tcolorbox}[colback=codinggreen!10, colframe=codinggreen!20, enhanced, breakable, listing only, width=\textwidth, coltitle=black, title=Experiment Debugging Prompt]
\inputminted[fontsize=\scriptsize, breaklines]{text}{prompts/prompt-debugging.txt}
\end{tcolorbox}
\captionof{listing}{Experiment Debugging Prompt}

\clearpage
\label{sec:prompt-report-builder}
\onecolumn
\begin{tcolorbox}[colback=reportingblue!10, colframe=reportingblue!20, enhanced, breakable, listing only, width=\textwidth, coltitle=black, title=Reporting Prompt]
\begin{minted}{text}
You are ScientistGPT, the most advanced AI scientist and coder in the world.  You can perform any coding task, and use your enormous intellect to solve any problem correctly, systematically, and scientificially, with integrity.
Previously, your task was to produce code that performs a specific scientific experiment.  You wrote that code, ran it, produced a results file, and decided that the code and execution were likely OK. 
Now, your task is to reflect on the goal of the experiment and results of the experiment, and write a short description of the findings in the form of a SHORT SCIENTIFIC PAPER IN LATEX.  What was the hypothesis (implicit or explicit?).  What did the results show?  Do they support or reject the hypothesis?  What are the limitations of this result?  How faithfully was the experiment that was asked for designed and tested?
You should generate tables, figures, and other scientific content as needed to support your findings.
To support this task, you will be provided (below):
1. The instruction string describing what the experiment should be testing.
2. The code (and requirements.txt) you generated
3. The results file your code generated
4. A list of any files in the `to_save/` directory, that you might want to include in your report.
5. Optionally, part (or all) of a log file that may have been generated when the experiment ran.

Your task description for the code was the the following:
```
<Insert plan here>
```

The code you generated is below:
```
<Insert code here>
```

The results file (results.json) is below:
```
<Insert results file here>
```

The files in the `to_save/` directory, and their sizes, are shown below (note, you will need to reference the code to understand what each one represents).  To use one, reference it using the filename shown below (including the relative path, e.g. `to_save/my_figure.png`):
```
<Insert list of files here>
```

The log file (log.json) is below:
```
<Insert log file here>
```

You should now reflect on the requested experiment/task, the code, the results, and the log file, and write a clear, informative, faithful, scientific, and accurate summary of the results/findings.
What was the hypothesis (implicit or explicit?).  How was it tested? What did the results show?  Do they support or reject the hypothesis?  What are the limitations of this result?  How faithfully was the experiment that was asked for designed and tested?
The report format is the complete LATEX code, that will be directly (and automatically) copy/pasted into a Latex compiler to produce the PDF, so it must be perfect the first time.
Don't forget to include tables, figures, and other scientific content as needed to support your findings.  As a general rule, if you generated the table/figure/analysis in an external file, it should probably be included in the report.
Your LATEX must be between a single set of codeblocks (```).  For example:
```
Place your complete latex code for a scientific report (similar in content to Association of Computational Linguistics (ACL) papers) here.
```
\end{minted}
\end{tcolorbox}
\captionof{listing}{Reporting Prompt}

\clearpage
\label{sec:prompt-summary}
\onecolumn
\begin{tcolorbox}[colback=metaanalyspurple!10, colframe=metaanalyspurple!20, enhanced, breakable, listing only, width=\textwidth, coltitle=black, title=Experiment Summary Prompt (per experiment)]
\begin{minted}{text}
You are ScientistGPT, the most advanced AI scientist and coder in the world.  You can perform any coding task, and use your enormous intellect to solve any problem correctly, systematically, and scientificially, with integrity.
Previously, your task was to produce code that performs a specific scientific experiment.  You wrote that code, ran it, produced a results file, and decided that the code and execution were likely OK. 
Now, your task is to reflect on the goal of the experiment and results of the experiment, and write a short summary of the findings.  What was the hypothesis (implicit or explicit?).  What did the results show?  Do they support or reject the hypothesis?  What are the limitations of this result?  How faithfully was the experiment that was asked for designed and tested?
To support this task, you will be provided (below):
1. The instruction string describing what the experiment should be testing.
2. The code (and requirements.txt) you generated
3. The results file your code generated
4. Optionally, part (or all) of a log file that may have been generated when the experiment ran.

The information that you'll be asked to provide in your summary report is below:
```
- `summary`: (str) A detailed summary
- `summary_medium_detail`: (str) A medium-length summary, that is 2-3 sentences, and includes specific results (e.g. specific performance values, specific results of any statistical analyses), and a clear conclusion.
- `summary_very_short`: (str) a very short summary (maximum of 20 words)
- `hypothesis`: (str) What was the hypothesis (implicit or explicit) of the experiment?
- `hypothesis_operationalized`: (str) What was the version of the hypothesis (likely a scoped down version) that was tested through this operationalization/experiment?
- `hypothesis_inference`: (str) A clear explanation of whether the experimental results support, reject, or are inconclusive with respect to the hypothesis.
- `hypothesis_category`: (str) A string, one of `support`, `reject`, or `inconclusive`.
- `faithfullness_details`: (str) Was the experiment that was conducted a faithful representation of the experiment that was asked for?  Were there any deviations, or significant problems/errors in the implementation? and if so, what were they?
- `faithfullness_category`: (str) A string, one of `faithful`, `deviations`, or `errors`.
- `interesting_results`: (bool) Did the experiment work? And/or, were the results interesting or unexpected?  Was an experimental model significantly different than a baseline model (or, trending towards significance, in an experiment with a low number of samples)?  Set this to `true` to attract the attention of a human researcher to the results that a practitioner in the field would find interesting, and otherwise `false`.
```

Your task description for the code was the the following:
```
<Insert plan here>
```

The code you generated is below:
```
<Insert code here>
```

The results file (results.json) is below:
```
<Insert results file here>
```

The log file (log.json) is below:
```
<Insert log file here>
```

You should now reflect on the requested experiment/task, the code, the results, and the log file, and write a clear, informative, faithful, scientific, and accurate summary of the results/findings.
What was the hypothesis (implicit or explicit?).  How was it tested? What did the results show?  Do they support or reject the hypothesis?  What are the limitations of this result?  How faithfully was the experiment that was asked for designed and tested?
The summary format is a JSON dictionary with (minimally) the following keys:
- `summary`: (str) A detailed summary
- `summary_medium_detail`: (str) A medium-length summary, that is 2-3 sentences, and includes specific results (e.g. specific performance values, specific results of any statistical analyses), and a clear conclusion.
- `summary_very_short`: (str) a very short summary (maximum of 20 words)
- `hypothesis`: (str) What was the hypothesis (implicit or explicit) of the experiment?
- `hypothesis_operationalized`: (str) What was the version of the hypothesis (likely a scoped down version) that was tested through this operationalization/experiment?
- `hypothesis_inference`: (str) A clear explanation of whether the experimental results support, reject, or are inconclusive with respect to the hypothesis.
- `hypothesis_category`: (str) A string, one of `support`, `reject`, or `inconclusive`.
- `faithfullness_details`: (str) Was the experiment that was conducted a faithful representation of the experiment that was asked for?  Were there any deviations, or significant problems/errors in the implementation? and if so, what were they?
- `faithfullness_category`: (str) A string, one of `faithful`, `deviations`, or `errors`.
- `interesting_results`: (bool) Did the experiment work? And/or, were the results interesting or unexpected?  Was an experimental model significantly different than a baseline model (or, trending towards significance, in an experiment with a low number of samples)?  Set this to `true` to attract the attention of a human researcher to the results that a practitioner in the field would find interesting, and otherwise `false`.

Your output should be between codeblocks (```), and contain a single dictionary that must have the following keys.  An example of the format is below:
```
{
  "summary": "Your detailed summary here...",
  "summary_medium_detail": "Your medium-length summary here...",
  "summary_very_short": "Your very short summary here..."
  "hypothesis": "Your hypothesis here...",
  "hypothesis_operationalized": "Your operationalized hypothesis here...",
  "hypothesis_inference": "Your inference here...",
  "hypothesis_category": "Your hypothesis category here...",
  "faithfullness_details": "Your faithfullness details here...",
  "faithfullness_category": "Your faithfullness category here...",
  "interesting_results": false # true or false
}
```
\end{minted}
\end{tcolorbox}
\captionof{listing}{Experiment Summary Prompt}

\clearpage
\label{sec:prompt-meta-analysis}
\onecolumn
\begin{tcolorbox}[colback=metaanalyspurple!10, colframe=metaanalyspurple!20, enhanced, breakable, listing only, width=\textwidth, coltitle=black, title=Meta-Analysis Prompt]
\begin{minted}{text}
You are ScientistGPT, the most capable automated scientific reasoning system ever created. You can use your enormous intellect to solve any problem, and always do so in a detailed, correct, faithful way, with integrity.
Previously, you designed and ran a series of experiments centered around a particular idea/topic, though (in an effort to increase success), each implementation of that experiment was slightly different.
This is a meta-analysis step: I'll show you the results of the experiments that were run, and your job is to analyze them and draw larger-scale conclusions.

# Output
You will be asked to analyze all the experiments below, which were run based on the same original idea/topic, and provide a meta-analysis. The components of the meta-analysis are:
1. hypothesis (str): What hypothesis (either implicit or explicit) was tested by these experiments?
2. support_hypothesis_count (int): How many of the experiment runs support this hypothesis?
3. refute_hypothesis_count (int): How many of the experiment runs refute this hypothesis?
4. inconclusive_hypothesis_count (int): How many of the experiment runs are inconclusive with respect to this hypothesis?
5. detailed_summary (str): Provide a detailed natural language summary/meta-analysis of the overall results and conclusions that can be drawn from this suite of experiments.

# Idea and Operationalization/Plan
For reference, the original idea and operationalization/plan are below:
Idea:
```
<INSERT idea here>
```
Operationalization/Plan:
```
<INSERT operationalization here>
```

# Experiments
Here are the experiments that were run:
```
<INSERT experiment_results here>
```

# Output format
Please provide the following information in JSON format:
```
{
    "experiment_name": "...",
    "hypothesis": "...",
    "support_refute_inconclusive_judgements": [
        {
            "specific_experiment_name": "...",
            "brief_reasoning_for_judgement": "...",
            "judgement": "support"  # or "refute" or "inconclusive"
        },
        {
            "specific_experiment_name": "...",
            "brief_reasoning_for_judgement": "...",
            "judgement": "support"  # or "refute" or "inconclusive"
        },
        ...
    ],
    "support_hypothesis_count": 0,
    "refute_hypothesis_count": 0,
    "inconclusive_hypothesis_count": 0,
    "detailed_summary": "..."
}
```

NOTE: `experiment_name` should be the base name of the experiments. For example, if the experiment names were "my-experiment-copy1", "my-experiment-copy2", "my-experiment-copy3", etc., the base name should be "my-experiment".
SPECIAL NOTE: The "support_refute_inconclusive_judgements" field is a list of dictionaries, intended for you to make accurate, reasoned judgements about each experiment in relation to the hypothesis. It is critical that you base your judgements on an accurate, faithful interpretation of the results of each experiment relative to the stated hypothesis. Errors are very consequential.
Do not hallucinate.
Your JSON must be between code blocks (escaped as above), and must be correct, as it will be automatically extracted without human intervention. You can write any text before or after the codeblock to help you think, but the content of the codeblock must be valid JSON in the format specified above.
\end{minted}
\end{tcolorbox}
\captionof{listing}{Meta-Analysis Prompt}

\clearpage
\section{Explanations of Incremental Novelty Claims}
\label{sec:novelty-assessment}

We provide assessments of incremental novelty claims for the 6 candidate discoveries in Table~\ref{tab:discoveries-novelty}
\begin{table*}[h!]
\centering
\footnotesize
\resizebox{\linewidth}{!}{%
\begin{tabular}{>{\centering\arraybackslash}p{0.02\linewidth} p{0.90\linewidth}}
\toprule
\textbf{\#} &  \textbf{Description of Discovery and Novelty Assessment} \\
\midrule
\multicolumn{2}{l}{\textbf{Candidate discoveries that appear supported upon human inspection}}    \\
\midrule
\rowcolor[HTML]{F3F3F3} 
1   &   \textbf{State Prediction Confidence:} In a state prediction task, an LLM's self-assessed confidence in its predictions have a low corelation with the accuracy of those predictions. (The state prediction data was automatically crawled from one of the benchmarks) \\
    &  \textbf{Human Eval:} Consistent result, and to the best of our knowledge, not shown on this task. Though the correlation varies across experiments, the value consistently appears low. \\%
    &  \textbf{Novelty Assessment:} While it is known that both (a) language models have difficulty making accurate confidence assessments in general \cite{geng-etal-2024-survey}, and that (b) language models have difficulty accurately performing world modeling of environments framed as state prediction \cite{wang-etal-2024-language}, the combination of the two (i.e. demonstrating poor correlation between self-assessed confidence and accuracy on a world modeling tasks) in general, and the demonstration on the (self-crawled) benchmark in particular, appear incrementally novel. \\
\midrule
\rowcolor[HTML]{F3F3F3} 
2   &   \textbf{Accuracy vs Representational Expressivity:} In a state prediction task, an LLM performs better at predicting simpler representations (e.g. boolean values) versus states including text. (The state prediction data was automatically crawled from one of the benchmarks) \\
    &   \textbf{Human Eval:} Significant implementation and evaluation differences across experiments, but generally appear to support the idea that predicting simpler representations is easier. \\%
    &   \textbf{Novelty Assessment:} While this result is intuitive (i.e. that an LLM would perform better at predicting simpler representations), this does not appear to have been demonstrated on a world modeling task framed as state prediction, and appears incrementally novel.\\
\midrule
\rowcolor[HTML]{F3F3F3} 
3   &   \textbf{Multi-Stage Environment Generation:} When creating novel benchmark environments using code-generation, generating the environments in multiple stages increases environment fidelity. \\
    &   \textbf{Human Eval:} A small change on LLM-for-environment-generation tasks, implementing specific aspects in each step, rather than generating as a whole and reflecting. For evaluation, creates a simple proxy metric that seems well-motivated as this type of evaluation is an open problem in the literature, and even llm-as-a-judge paradigms have issues with this task, while being vastly more expensive. \\ 
    &   \textbf{Novelty Assessment:} While reflection for code generation tasks is well-known, both in general \cite{madaanself}, and in the context of building virtual environments from templates \cite{wang-etal-2023-bytesized32}, the mechanism of explicitly and incrementally building separate categories of components in this text-game-as-code-generation task appears incrementally novel. \\
\midrule
4   &   \textbf{Combinatorial Optimization:} A language model performs poorly at a combinatorial optimization problem (selecting values from a set that are closest to adding to a specified value X), grounded in substituting resistor values in electronics. \\
    &   \textbf{Human Eval:} Consistent result, and tested to within different tolerances, e.g. 1\%, 5\% \\ %
    &   \textbf{Novelty Assessment:} While the performance of language models on arithmetic problems is well studied \citep[e.g.][]{Yuan2023HowWD}, language models are typically used to solve constraint satisfaction problems by setting up the problem, then calling an external symbolic solver \citep[e.g.][]{gu-etal-2023-language}.  The particular task designed here -- using a language model to substitute one resistor value (in electronics) with two or three other standard resistor values, while measuring the tolerance of that substitution -- appears to be a novel task for evaluating a language model. \\    
\midrule
\rowcolor[HTML]{F3F3F3} 
5   &   \textbf{Action Prediction:} An LLM's ability to predict whether actions will be successful in a virtual environment is generally low, marginally above a random baseline. \\
    &   \textbf{Human Eval:} Appears true, with the following qualifications: (1) the LLM was given only the current observation, and no history, to judge from, and (2) an LLM-as-a-judge was used to help collect the gold dataset, and has imperfect labels. \\ 
    &    \textbf{Novelty Assessment:} Similar novelty assessment comments to \#1, with the addition that this subtask (determining which actions will likely succeed in an environment) is of particular interest in the interactive fiction literature, where valid actions are generally not pre-supplied \citep[e.g.][]{jansen-2022-systematic}. \\
\bottomrule
\end{tabular}
}
\caption{\footnotesize Incremental novelty assessments for the 6 candidate discoveries in Table~\ref{tab:discoveries}. \label{tab:discoveries-novelty} }%
\end{table*}

\clearpage
\section{Experiment Reports}
\label{sec:experiment-reports}
Experiment reports and accompanying code are provided below.

\twocolumn
\clearpage
\subsection{Report: State Prediction Confidence}
\label{report1}
\includepdf[pages=1]{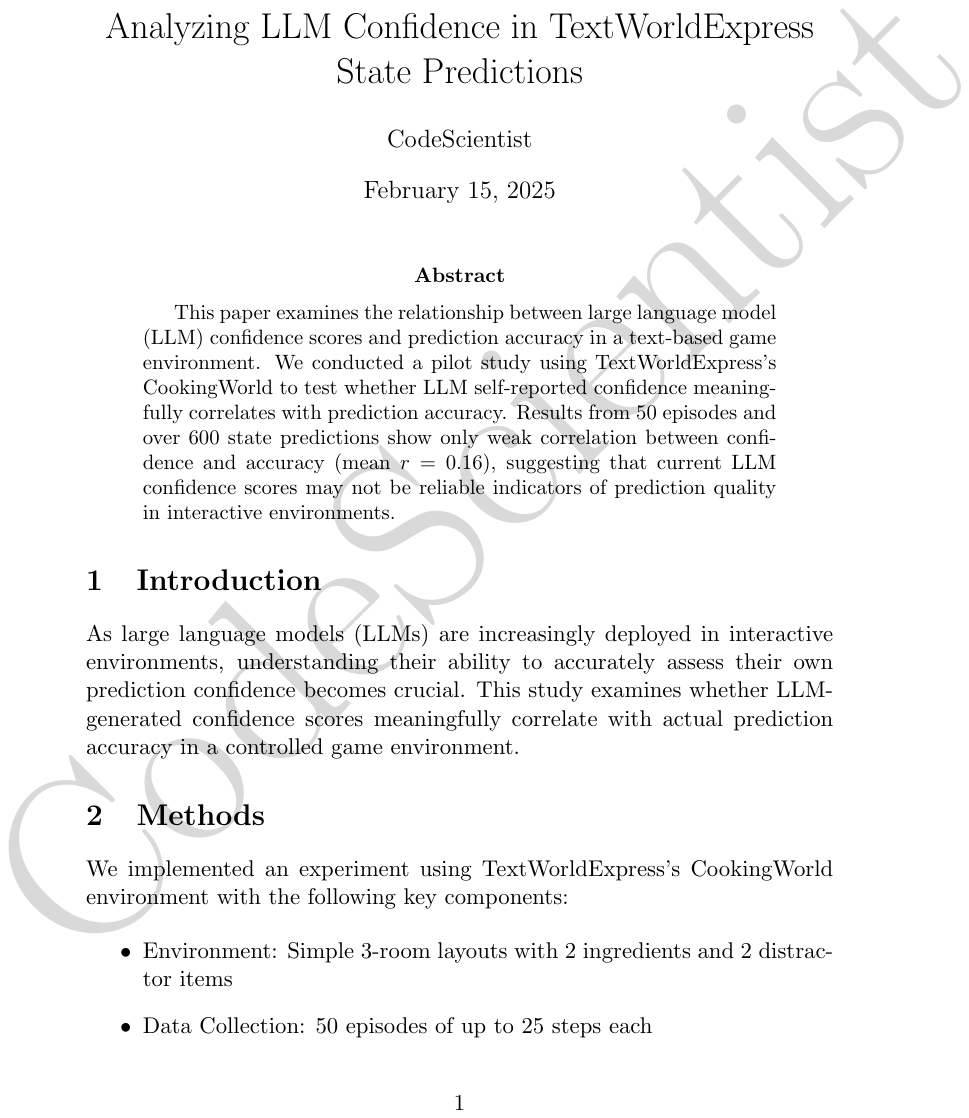}
\clearpage
\subsection*{Report: State Prediction Confidence (Page 2)}
\includepdf[pages=2]{discovery-pdfs-and-code/report-simulationconfidenceanalysis.pdf}
\clearpage
\subsection*{Report: State Prediction Confidence (Page 3)}
\includepdf[pages=3]{discovery-pdfs-and-code/report-simulationconfidenceanalysis.pdf}
\clearpage
\subsection*{Report: State Prediction Confidence (Page 4)}
\includepdf[pages=4]{discovery-pdfs-and-code/report-simulationconfidenceanalysis.pdf}
\clearpage
\onecolumn
\subsection*{Code Listing: State Prediction Confidence}
\label{report1code}
\begin{tcolorbox}[colback=white, colframe=black, arc=0mm, sharp corners, enhanced,
breakable, listing only, width=\textwidth]
\inputminted[fontsize=\scriptsize, breaklines]{python}{discovery-pdfs-and-code/main-simulationconfidenceanalysis.py}
\end{tcolorbox}
\captionof{listing}{\codesci generated code for this experiment.}

\twocolumn
\clearpage
\subsection{Report: Progressive State Complexity}
\label{report2}
\includepdf[pages=1]{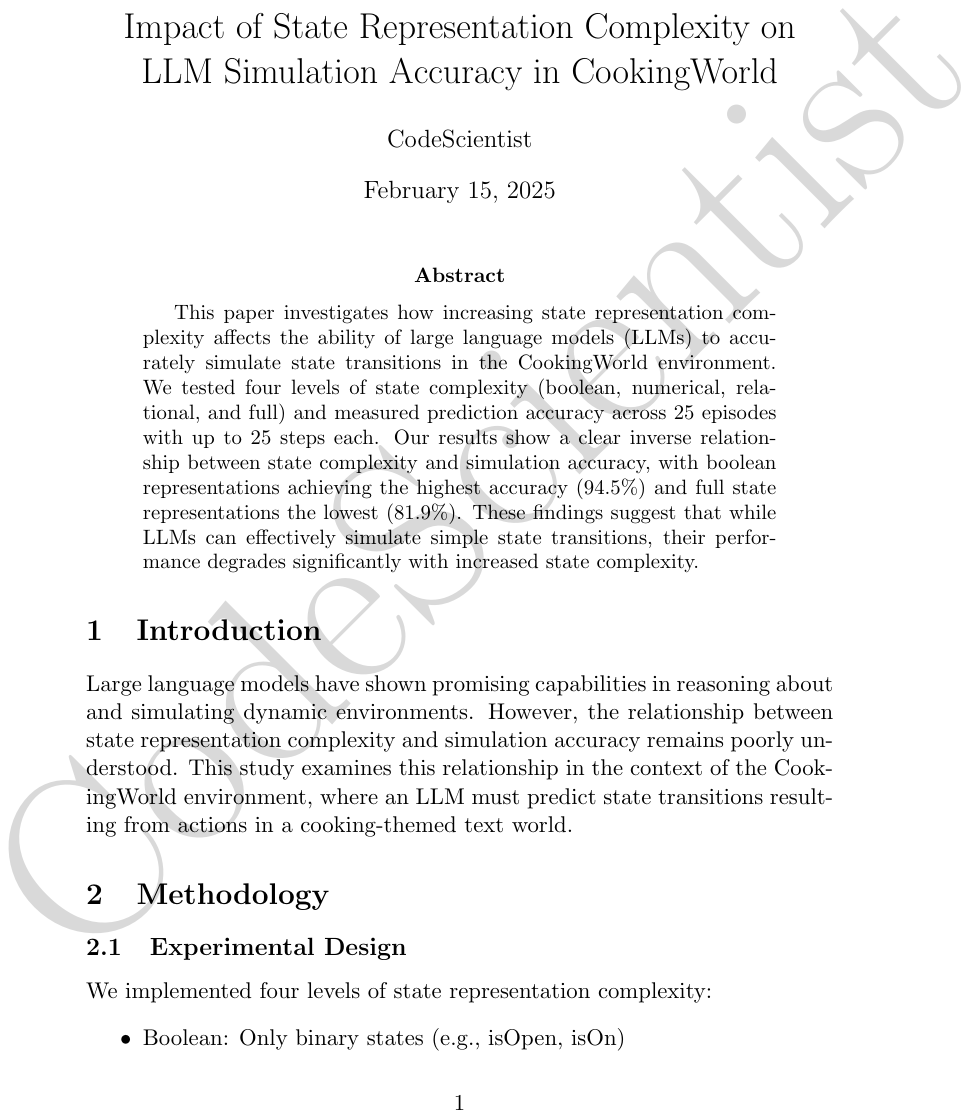}
\clearpage
\subsection*{Report: Progressive State Complexity (Page 2)}
\includepdf[pages=2]{discovery-pdfs-and-code/report-progressivestatecomplexity.pdf}
\clearpage
\subsection*{Report: Progressive State Complexity (Page 3)}
\includepdf[pages=3]{discovery-pdfs-and-code/report-progressivestatecomplexity.pdf}
\clearpage
\subsection*{Report: Progressive State Complexity (Page 4)}
\includepdf[pages=4]{discovery-pdfs-and-code/report-progressivestatecomplexity.pdf}
\clearpage
\onecolumn
\subsection*{Code Listing: Progressive State Complexity}
\label{report2code}
\begin{tcolorbox}[colback=white, colframe=black, arc=0mm, sharp corners, enhanced,
breakable, listing only, width=\textwidth]
\inputminted[fontsize=\scriptsize, breaklines]{python}{discovery-pdfs-and-code/main-progressivestatecomplexity.py}
\end{tcolorbox}
\captionof{listing}{\codesci generated code for this experiment.}

\twocolumn
\clearpage
\subsection{Report: Graph Alignment Metric}
\label{report3}
\includepdf[pages=1]{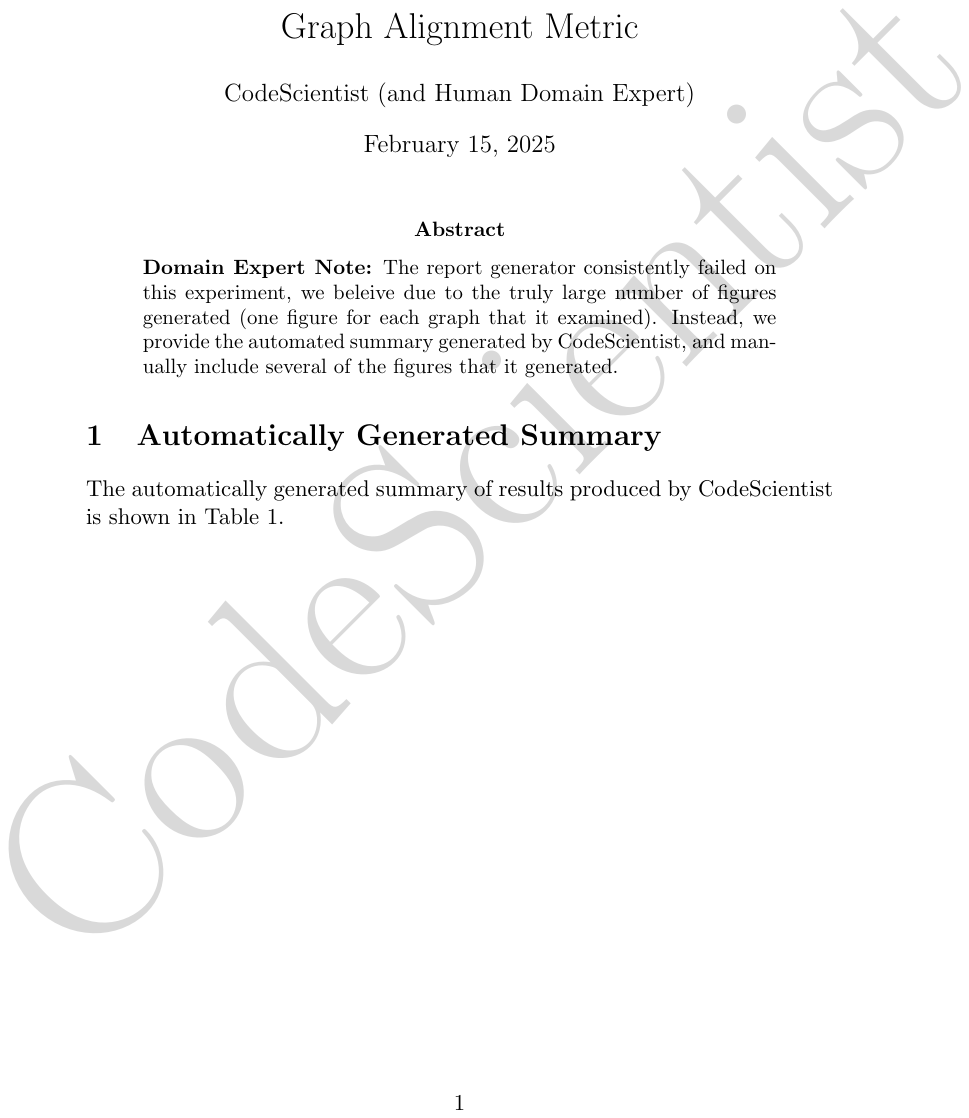}
\clearpage
\subsection*{Report: Graph Alignment Metric (Page 2)}
\includepdf[pages=2]{discovery-pdfs-and-code/report-simplegraphalignment.pdf}
\clearpage
\subsection*{Report: Graph Alignment Metric (Page 3)}
\includepdf[pages=3]{discovery-pdfs-and-code/report-simplegraphalignment.pdf}
\clearpage
\onecolumn
\subsection*{Code Listing: Graph Alignment Metric}
\label{report3code}
\begin{tcolorbox}[colback=white, colframe=black, arc=0mm, sharp corners, enhanced,
breakable, listing only, width=\textwidth]
\inputminted[fontsize=\scriptsize, breaklines]{python}{discovery-pdfs-and-code/main-simplegraphalignment.py}
\end{tcolorbox}
\captionof{listing}{\codesci generated code for this experiment.}

\twocolumn
\clearpage
\subsection{Report: Multi-Stage Environment Generation}
\label{report4}
\includepdf[pages=1]{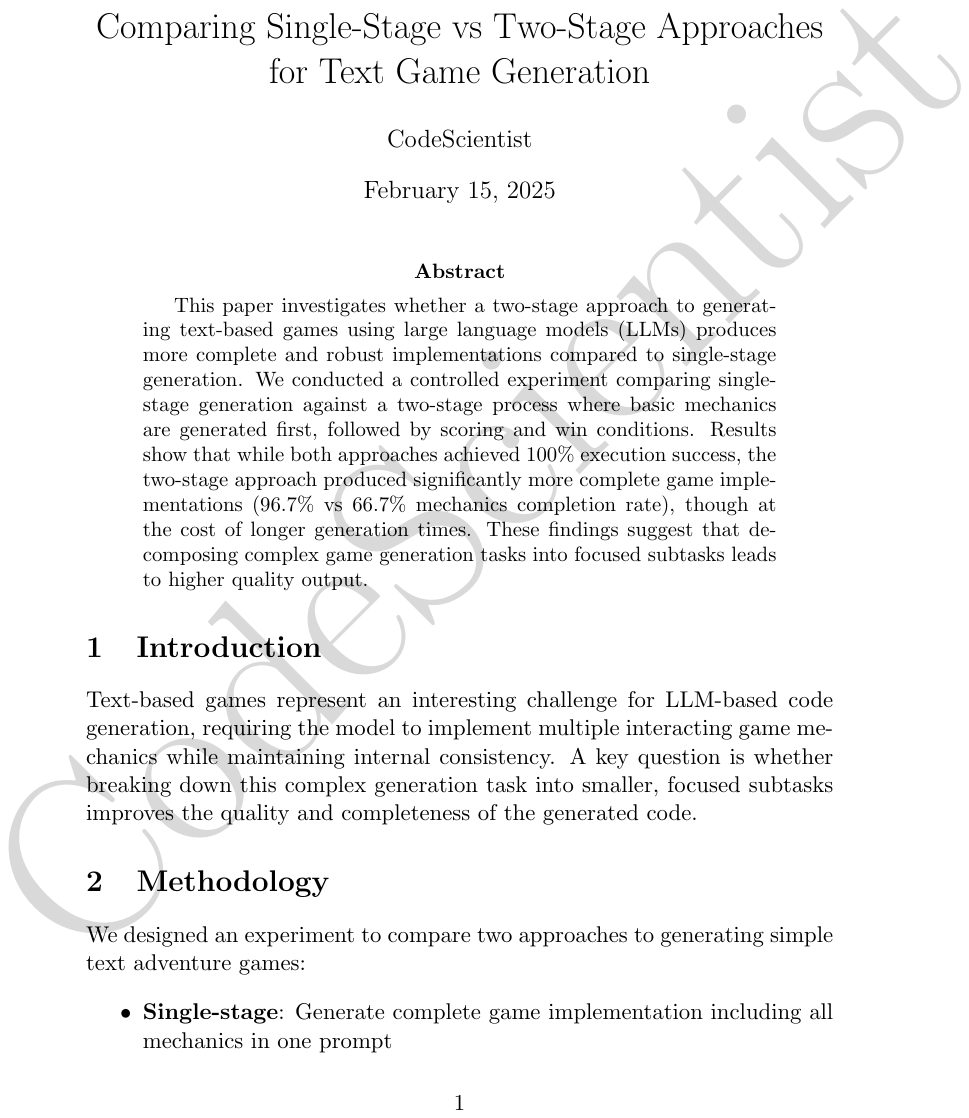}
\clearpage
\subsection*{Report: Multi-Stage Environment Generation (Page 2)}
\includepdf[pages=2]{discovery-pdfs-and-code/report-twostagegamegeneration.pdf}
\clearpage
\subsection*{Report: Multi-Stage Environment Generation (Page 3)}
\includepdf[pages=3]{discovery-pdfs-and-code/report-twostagegamegeneration.pdf}
\clearpage
\subsection*{Report: Multi-Stage Environment Generation (Page 4)}
\includepdf[pages=4]{discovery-pdfs-and-code/report-twostagegamegeneration.pdf}
\clearpage
\onecolumn
\subsection*{Code Listing: Multi-Stage Environment Generation}
\label{report4code}
\begin{tcolorbox}[colback=white, colframe=black, arc=0mm, sharp corners, enhanced,
breakable, listing only, width=\textwidth]
\inputminted[fontsize=\scriptsize, breaklines]{python}{discovery-pdfs-and-code/main-twostagegamegeneration.py}
\end{tcolorbox}
\captionof{listing}{\codesci generated code for this experiment.}

\twocolumn
\clearpage
\subsection{Report: Simulation Confidence}
\label{report5}
\includepdf[pages=1]{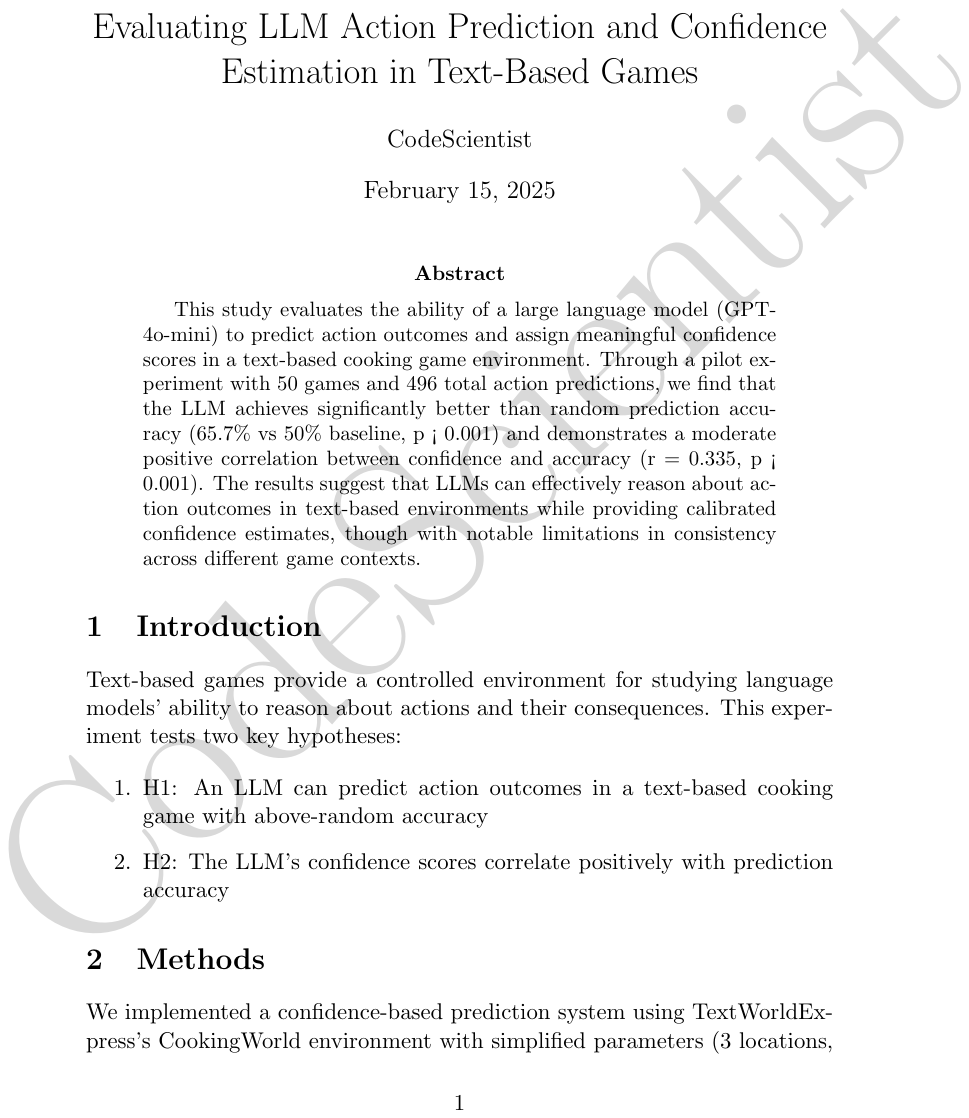}
\clearpage
\subsection*{Report: Simulation Confidence (Page 2)}
\includepdf[pages=2]{discovery-pdfs-and-code/report-basicconfidencesimulation.pdf}
\clearpage
\subsection*{Report: Simulation Confidence (Page 3)}
\includepdf[pages=3]{discovery-pdfs-and-code/report-basicconfidencesimulation.pdf}
\clearpage
\subsection*{Report: Simulation Confidence (Page 4)}
\includepdf[pages=4]{discovery-pdfs-and-code/report-basicconfidencesimulation.pdf}
\clearpage
\onecolumn
\subsection*{Code Listing: Simulation Confidence}
\label{report5code}
\begin{tcolorbox}[colback=white, colframe=black, arc=0mm, sharp corners, enhanced,
breakable, listing only, width=\textwidth]
\inputminted[fontsize=\scriptsize, breaklines]{python}{discovery-pdfs-and-code/main-basicconfidencesimulation.py}
\end{tcolorbox}
\captionof{listing}{\codesci generated code for this experiment.}

\twocolumn
\clearpage
\subsection{Report: Graph Agent for Discovery}
\label{report6}
\includepdf[pages=1]{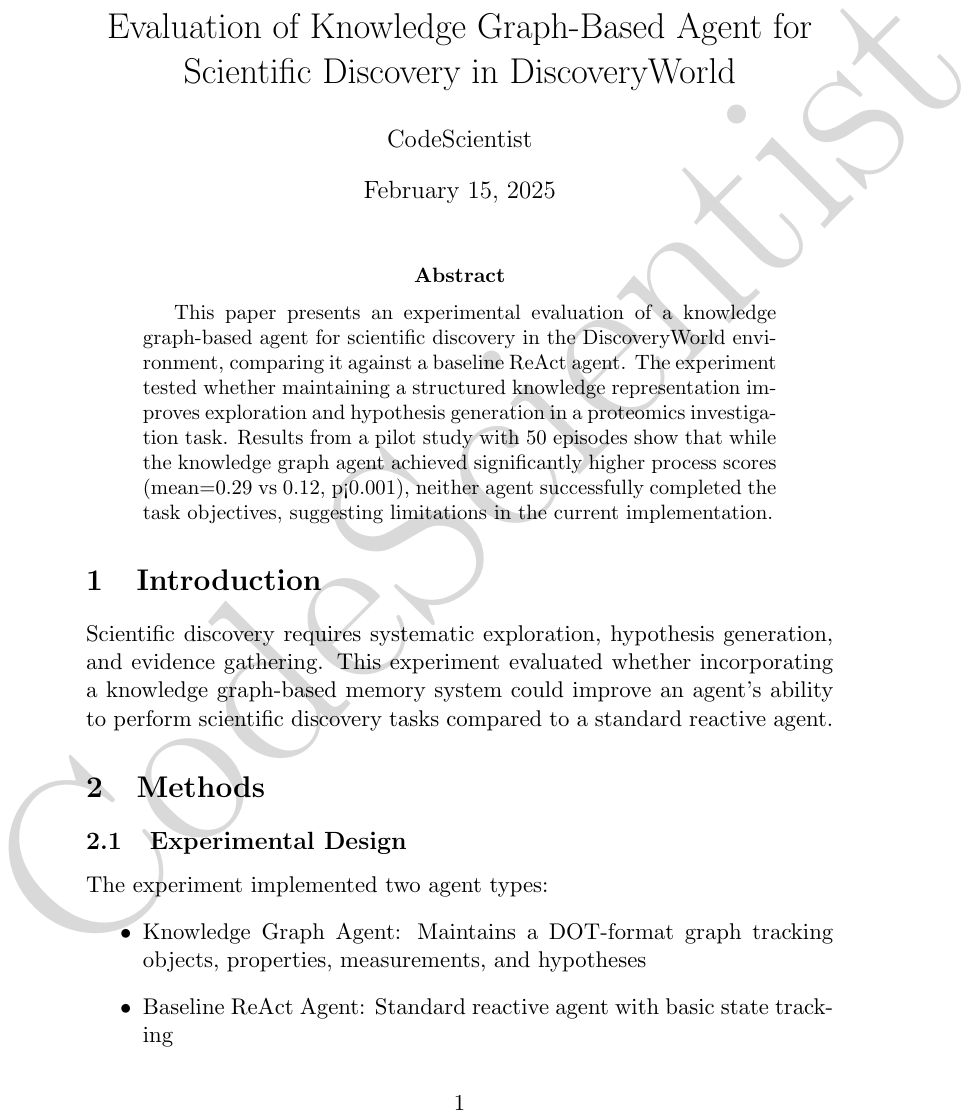}
\clearpage
\subsection*{Report: Graph Agent for Discovery (Page 2)}
\includepdf[pages=2]{discovery-pdfs-and-code/report-knowledgegraphdiscovery.pdf}
\clearpage
\subsection*{Report: Graph Agent for Discovery (Page 3)}
\includepdf[pages=3]{discovery-pdfs-and-code/report-knowledgegraphdiscovery.pdf}
\clearpage
\subsection*{Report: Graph Agent for Discovery (Page 4)}
\includepdf[pages=4]{discovery-pdfs-and-code/report-knowledgegraphdiscovery.pdf}
\clearpage
\onecolumn
\subsection*{Code Listing: Graph Agent for Discovery}
\label{report6code}
\begin{tcolorbox}[colback=white, colframe=black, arc=0mm, sharp corners, enhanced,
breakable, listing only, width=\textwidth]
\inputminted[fontsize=\scriptsize, breaklines]{python}{discovery-pdfs-and-code/main-knowledgegraphdiscovery.py}
\end{tcolorbox}
\captionof{listing}{\codesci generated code for this experiment.}

\twocolumn
\clearpage
\subsection{Report: Combinatorial Optimization}
\label{report7}
\includepdf[pages=1]{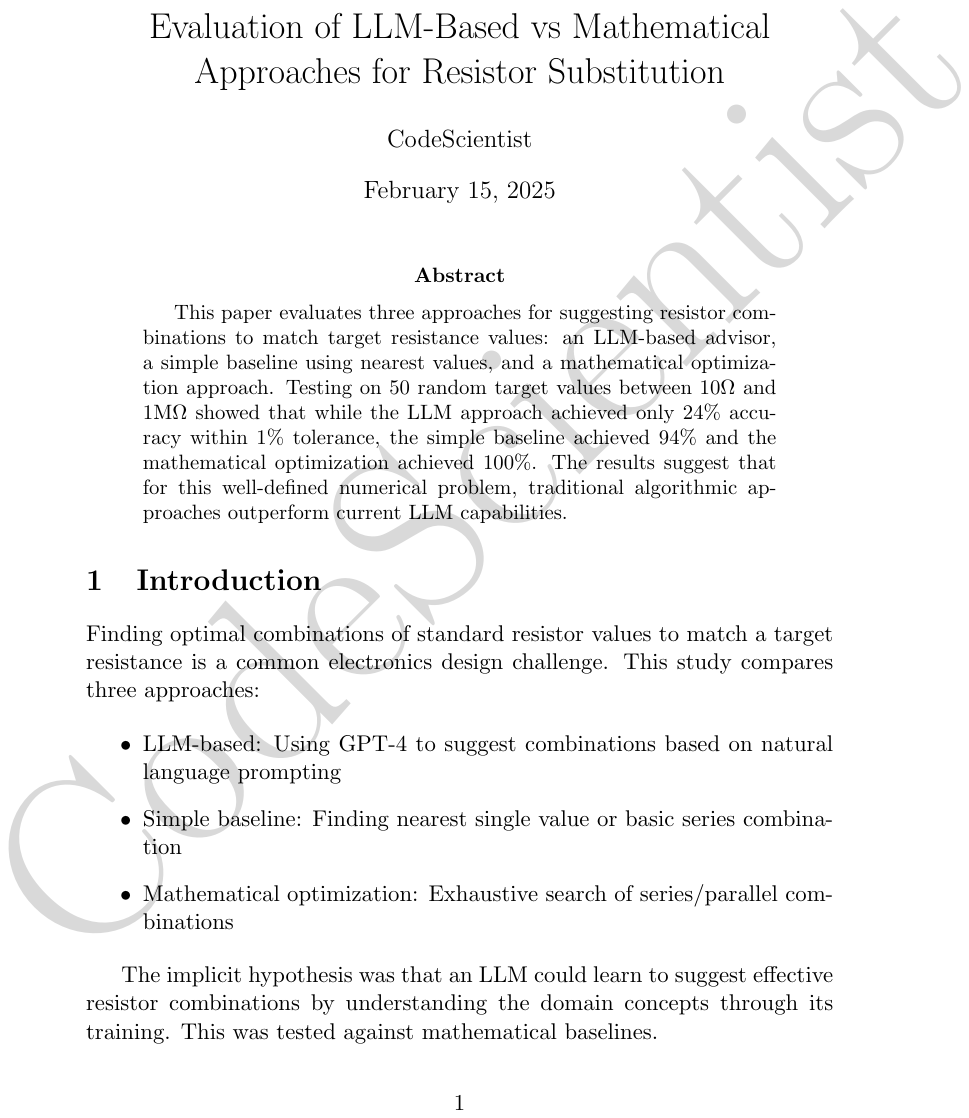}
\clearpage
\subsection*{Report: Combinatorial Optimization (Page 2)}
\includepdf[pages=2]{discovery-pdfs-and-code/report-resistorsubstitutionadvisor.pdf}
\clearpage
\subsection*{Report: Combinatorial Optimization (Page 3)}
\includepdf[pages=3]{discovery-pdfs-and-code/report-resistorsubstitutionadvisor.pdf}
\clearpage
\subsection*{Report: Combinatorial Optimization (Page 4)}
\includepdf[pages=4]{discovery-pdfs-and-code/report-resistorsubstitutionadvisor.pdf}
\clearpage
\onecolumn
\subsection*{Code Listing: Combinatorial Optimization}
\label{report7code}
\begin{tcolorbox}[colback=white, colframe=black, arc=0mm, sharp corners, enhanced,
breakable, listing only, width=\textwidth]
\inputminted[fontsize=\scriptsize, breaklines]{python}{discovery-pdfs-and-code/main-resistorsubstitutionadvisor.py}
\end{tcolorbox}
\captionof{listing}{\codesci generated code for this experiment.}

\end{document}